\pdfoutput=1
\documentclass[manuscript, screen=True]{acmart}

\usepackage{xcolor}
\usepackage{multirow}

\AtBeginDocument{%
  }

\setcopyright{acmcopyright}
\copyrightyear{2023}
\acmYear{2023}
\acmDOI{XXXXXXX.XXXXXXX}

\acmJournal{CSUR}
\acmVolume{37}
\acmNumber{4}
\acmArticle{111}
\acmMonth{8}

\begin{document}

\title{A Survey on Interpretable Cross-modal Reasoning}

\author{Dizhan Xue}
\email{xuedizhan17@mails.ucas.ac.cn}
\orcid{0000-0002-0173-1556}

\author{Shengsheng Qian}
\email{shengsheng.qian@nlpr.ia.ac.cn}
\orcid{0000-0001-9488-2208}

\author{Zuyi Zhou}
\email{zhouzuyi2023@ia.ac.cn}
\orcid{0009-0008-0306-8461}

\affiliation{%
  \institution{MAIS, Institute of Automation, Chinese Academy of Sciences}
  \streetaddress{95 Zhongguancun East Road}
  \city{Beijing}
  \state{Beijing}
  \country{China}
  \postcode{100190}
}
\affiliation{%
  \institution{School of Artificial Intelligence, University of Chinese Academy of Sciences}
  \streetaddress{No.1 Yanqihu East Rd, Huairou District}
  \city{Beijing}
  \state{Beijing}
  \country{China}
  \postcode{101408}
}

\author{Changsheng Xu}
\email{csxu@nlpr.ia.ac.cn}
\orcid{0000-0001-8343-9665}
\authornote{Corresponding author}

\affiliation{%
  \institution{MAIS, Institute of Automation, Chinese Academy of Sciences}
  \streetaddress{95 Zhongguancun East Road}
  \city{Beijing}
  \state{Beijing}
  \country{China}
  \postcode{100190}
}
\affiliation{%
  \institution{School of Artificial Intelligence, University of Chinese Academy of Sciences}
  \streetaddress{No.1 Yanqihu East Rd, Huairou District}
  \city{Beijing}
  \state{Beijing}
  \country{China}
  \postcode{101408}
}
\affiliation{%
  \institution{Peng Cheng Laboratory}
  \streetaddress{No.2 Xingke 1st Street, Nanshan District}
  \city{Shenzhen}
  \state{Guangdong}
  \country{China}
  \postcode{518066}
}

\renewcommand{\shortauthors}{Xue et al.}

\begin{abstract}
In recent years, cross-modal reasoning (CMR), the process of understanding and reasoning across different modalities, has emerged as a pivotal area with applications spanning from multimedia analysis to healthcare diagnostics.
As the deployment of AI systems becomes more ubiquitous, the demand for transparency and comprehensibility in these systems' decision-making processes has intensified. 
This survey delves into the realm of interpretable cross-modal reasoning (I-CMR), where the objective is not only to achieve high predictive performance but also to provide human-understandable explanations for the results.
This survey presents a comprehensive overview of the typical methods with a three-level taxonomy for I-CMR.
Furthermore, this survey reviews the existing CMR datasets with annotations for explanations.
Finally, this survey summarizes the challenges for I-CMR and discusses potential future directions.
In conclusion, this survey aims to catalyze the progress of this emerging research area by providing researchers with a panoramic and comprehensive perspective, illuminating the state of the art and discerning the opportunities.
The summarized methods, datasets, and other resources are available at \href{https://github.com/ZuyiZhou/Awesome-Interpretable-Cross-modal-Reasoning}{https://github.com/ZuyiZhou/Awesome-Interpretable-Cross-modal-Reasoning}.
\end{abstract}



\received{August 2023}
\received[revised]{XXXX}
\received[accepted]{XXXX}

\maketitle

\section{Introduction}
In recent years, there has been a growing interest in exploring the synergies between different modalities, such as vision and language, to enable machines to comprehend complex scenes, objects, and concepts in a more holistic manner \cite{baltruvsaitis2018multimodal, suzuki2022survey, xu2023multimodal}.
In particular, cross-modal reasoning (CMR) \cite{malkinski2022review, kaur2021comparative, sampat2022reasoning} refers to the process of understanding and reasoning across different modalities, such as image, text, audio, and video. 
It involves leveraging the relationships and interactions between multiple modalities to extract meaningful information and draw inferences.
CMR has given rise to numerous practical applications and tasks, including but not limited to visual question answering \cite{antol2015vqa, hudson2019gqa}, cross-modal retrieval \cite{qian2021dual, qian2022integrating}, vision-and-language navigation \cite{chen2019touchdown, majumdar2020improving}, visual grounding \cite{deng2018visual, huang2022deconfounded}, image-guided story generation \cite{xue2022mmt}.
Traditionally, most AI systems have focused on processing and understanding data within a single modality, such as analyzing images or text. However, in many real-world scenarios, information is available in multiple modalities simultaneously. Cross-modal reasoning aims to bridge the gap between these modalities and enable machines to reason and make decisions by integrating information from different sources.

More recently, the academic community has realized
the importance of interpretability in cross-modal reasoning \cite{HE2021104194, chen2022rex, lin2023zero}.
Interpretable cross-modal reasoning (I-CMR) refers to the ability to provide human-understandable explanations or justifications for the reasoning process behind a cross-modal reasoning model's predictions or decisions. This interpretability aspect not only enhances the transparency and trustworthiness of the model but also enables users to gain deeper insights into the underlying mechanisms driving cross-modal reasoning.
While recent cross-modal reasoning models have demonstrated their efficacy in various tasks, such as image captioning, visual question answering, and multimodal sentiment analysis, the black-box nature \cite{liang2021explaining, gaur2021semantics} of many cross-modal models limit their interpretability and applicability. 
As the complexity of these models increases, the need for I-CMR becomes ever more crucial.

Methods for I-CMR can be roughly classified into five categories according to the modality of the provided explanation as follows:
(1) Visual explanation \cite{lyu2022dime, liang2023multiviz} visualizes the contribution and impact of fine-grained multimodal features in CMR tasks or the relationships between these features across different modalities.
(2) Textual explanation \cite{tseng2022relation,zhang2023multimodal} elucidates the logical reasoning process or presents substantiating evidence for CMR in the form of text.
(3) Graph explanation \cite{zhang2022query, ding2022mukea} constructs graphs to demonstrate the extracted entities and relations for CMR or further highlight the reasoning path in the graph.
(4) Symbol explanation \cite{liu2023interpretable, gupta2023visual} symbolizes and facilitates the operation of the CMR process by logical inference proofs or the combination of atom programs.
(5) Multimodal explanation \cite{chen2022rex, yao2023beyond} aims to provide more comprehensive or user-friendly explanations by combining independent explanations of different modalities or generating a joint multimodal explanation.

\begin{figure}[t]
	\includegraphics[width=0.5\linewidth]{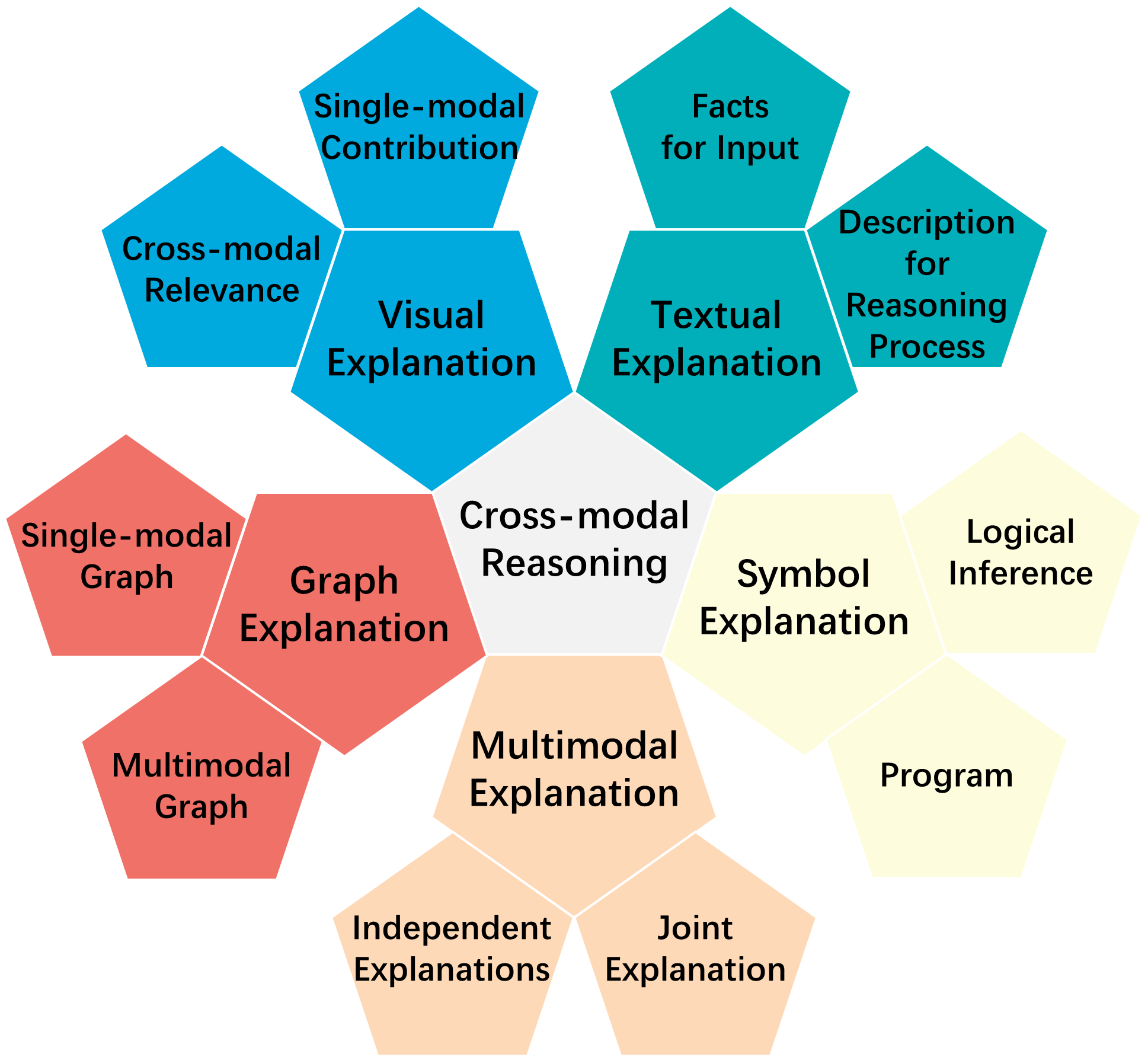}
		\centering
	\caption{Taxonomy for interpretable cross-modal reasoning (I-CMR).}
	\label{fig:tax}
\end{figure}

To the best of our knowledge, this is the first survey on interpretable cross-modal reasoning.
While some published surveys have focused on other topics in explainable machine learning, such as explainable AI and interpretable visual reasoning, we particularly focus on interpretable cross-modal reasoning in this survey.
We summarize some related surveys as follows: He et al. \cite{HE2021104194} reviewed methods for interpretable visual reasoning and associating datasets published in 2020 and before.
Gilpin et al. \cite{gilpin2018explaining} and Do{\v{s}}ilovi{\'c} et al. \cite{dovsilovic2018explainable} reviewed work in explanatory artificial intelligence published in 2018 and before.
Schoenborn et al. \cite{schoenborn2021explainable} reviewed explainable case-based reasoning approaches published in 2020 and before.
Recently, Dwivedi et al. \cite{dwivedi2023explainable} reviewed core ideas, techniques, and solutions in explainable AI.
%
%
Different from the above surveys, we review extensive research on interpretable cross-modal reasoning published in 2023Q2 and before to provide a comprehensive and up-to-date survey.

\begin{figure*}[t]
	\includegraphics[width=1\linewidth]{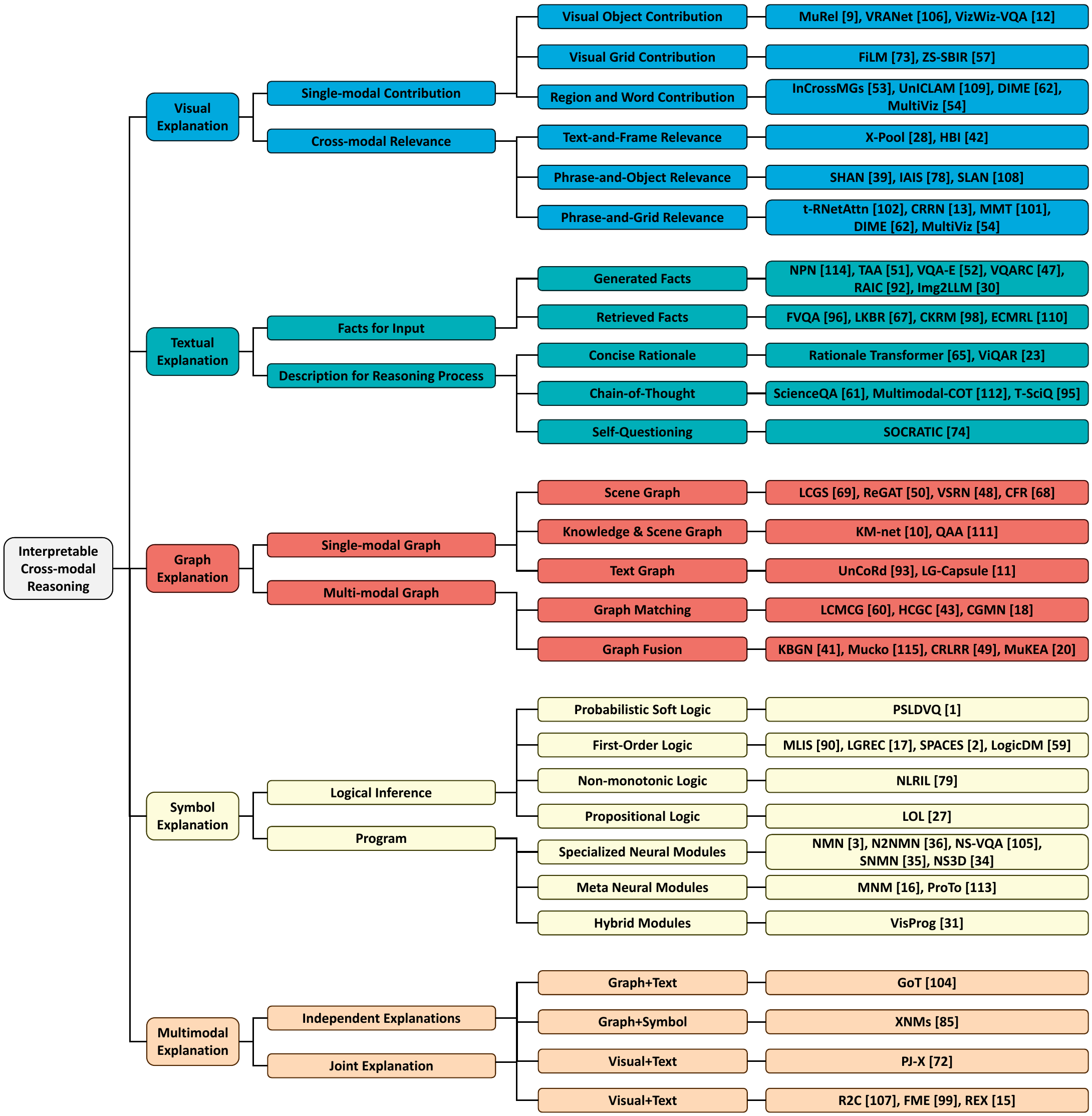}
		\centering
	\caption{Hierarchical classification of interpretable cross-modal reasoning methods.}
	\label{fig:overall}
\end{figure*}

The goal of this survey is to provide a comprehensive overview of the current state-of-the-art in interpretable cross-modal reasoning (I-CMR). Our contribution can be summarized as follows:
\begin{itemize}
    \item \textbf{New Taxonomy} We propose a three-level hierarchical taxonomy to classify existing work for I-CMR, as shown in Fig. \ref{fig:overall}. At the first level, current methods can be categorized into five groups: visual explanation, textual explanation, graph explanation, symbol explanation, and multimodal explanation.

    \item \textbf{Comprehensive review} We review various I-CMR models with associating explaining approaches. By surveying the previous literature, we will examine the key concepts, similarities, and limitations associated with existing I-CMR methods, shedding light on the current landscape of I-CMR.

    \item \textbf{Abundant resources} We collect and discuss abundant resources on I-CMR, including practical tasks, state-of-the-art methods, and public datasets. We hope this survey can serve as a valuable resource for anyone interested in the intersection of interpretable reasoning and cross-modal reasoning.

    \item \textbf{Challenges and future directions} After reviewing the state of the art of I-CMR, we summarize and analyze the existing challenges and future directions. We aim to provide insights into the latest developments and guide future research efforts toward building more advanced I-CMR models.
\end{itemize}

%
%

\section{Taxonomy for Interpretable Cross-modal Reasoning}
Fig. \ref{fig:overall} summarizes the general taxonomy of interpretable cross-modal reasoning (I-CMR) and the related methods reviewed in this paper.
The taxonomy is inspired by the modalities and design details of the provided explanations for CMR.
To begin with, according to the modality of the explanations, methods for I-CMR can be classified into five categories: visual explanation, textual explanation, graph explanation, symbol explanation, and multimodal explanation.
We further consider the design details of different methods and classify them in a hierarchical manner for ease of understanding and comparative analysis.

\subsection{Taxonomy of Visual Explanation}
This type of explanation utilizes visualization techniques to intuitively analyze and demonstrate the CMR process.
The methods of visual explanation can be further classified into two categories: Single-modal Contribution and Cross-modal Relevance.

\textbf{Single-modal Contribution}\quad These methods aim to visualize the contribution of different parts in the single-modal input. According to the particular definition of the input part, these methods can be divided into three groups: (1) \textbf{Visual Object Contribution}: These methods extract visual objects in the visual input (e.g., image and video) and attach importance scores to specific objects. The importance scores are visualized to show the contribution of different objects. (2) \textbf{Visual Grid Contribution}: These methods attach importance scores to different grids in the input image. The scores are visualized to show the contribution of different grids. (3) \textbf{Region and Word Contribution}: Besides visualizing the contribution of visual regions, these methods meanwhile compute and visualize the importance scores of different words in the textual input.

\textbf{Cross-modal Relevance}\quad These methods aim to visualize the relevance between multimodal inputs computed by the reasoning model. According to the specific parts in different modalities between which the relevance is computed, these methods can be categorized into three classes: (1) \textbf{Text-and-Frame Relevance}: These methods compute and visualize the relevance between texts and video frames. (2) \textbf{Phrase-and-Object Relevance}: These methods focus on the relevance between phrases in the textual input and the extracted objects in the visual input. (3) \textbf{Phrase-and-Grid Relevance}: These methods visualize the relevance between phrases in the textual input and grids in the visual input.

\subsection{Taxonomy of Textual Explanation}
This type of explanation interprets CMR processes in the form of legible text.
The methods of textual explanation can be further classified into two categories: Facts for Input and Description for Reasoning Process.

\textbf{Facts for Input}\quad The target of these methods target is to provide facts for the input to explain the CMR process. According to the manners by which the facts are obtained, these methods can be classified into two categories: (1) \textbf{Generated Facts}: These methods integrate generative modules and generate facts for the input, such as caption for the image. (2) \textbf{Retrieved Facts}: These methods retrieve facts from external knowledge bases, such as the relationship between key entities in the multimodal input.

\textbf{Description for Reasoning Process}\quad These methods attempt to generate textual descriptions for the reasoning process. According to the specific forms of the description, these methods are divided into three groups: (1) \textbf{Concise Rationale}: These methods attempt to concisely describe the reasoning process, typically in a single sentence. (2) \textbf{Chain-of-Thought}: These methods comprehensively describe the reasoning process step-by-step in the form of chain-of-thought. (3) \textbf{Self-Question}: These methods decompose the reasoning process into multiple sub-questions and backtrack all sub-questions and associating answers.

\subsection{Taxonomy of Graph Explanation}
This type of explanation constructs graphs to represent the entities and their relations to facilitate the explanation of the reasoning process.
The methods of graph explanation can be further classified into two categories: Single-modal Graph and Multi-modal graph.

\textbf{Single-modal Graph}\quad These methods construct the graph based on the single-modal input. According to the type of constructed graphs, these methods are categorized into three classes: (1) \textbf{Scene Graph}: These methods construct scene graphs based on the visual input, where nodes are visual concepts in the input. (2) \textbf{Knowledge \& Scene Graph}: These methods further expand the extracted scene graph by retrieving related entities and relations from external knowledge graphs. (3) \textbf{Text Graph}: These methods construct graphs based on the textual input, where nodes represent phrases or words.

\textbf{Multi-modal Graph}\quad These methods construct graphs based on the multimodal input. According to the interactions between graphs of different modalities, these methods are divided into two groups: (1) \textbf{Graph Matching}: These methods conduct the matching of graph structure to compute the similarity or relevance between nodes in different graphs. (2) \textbf{Graph Fusion} These methods fuse graphs of different modalities to perform comprehensive reasoning for the multimodal input.

\subsection{Taxonomy of Symbol Explanation}
This type of explanation symbolizes the CMR process and provides a symbol deduction of the reasoning result.
The methods of symbol explanation can be further classified into two categories: Logical Inference and Program.

\textbf{Logical Inference}\quad These methods conduct logical inferences to deduce the reasoning result. According to the adopted formal systems, these methods can be grouped into four classes: (1) \textbf{Probabilistic Soft Logic}: These methods adopt probabilistic soft logic (PSL) for inference. (2) \textbf{First-Order Logic}: These methods adopt first-order logic (FOL) for inference. (3) \textbf{Non-monotonic Logic}: These methods adopt non-monotonic logic (NML) for inference. (4) \textbf{Propositional Logic}: These methods adopt propositional logic (PL) for inference.

\textbf{Program}\quad These methods generate multiple modular implementations for program execution. According to the type of program modules, these methods are classified into three categories: (1) \textbf{Specialized Neural Modules}: These methods parameterize multiple neural modules to implement specialized functions. (2) \textbf{Meta Neural Modules}: These methods parameterize a single meta neural module to implement all involved functions. (3) \textbf{Hybrid Modules}: These methods adopt both neural modules and programming modules (e.g., Python functions) to implement various functions.

\subsection{Taxonomy of Multimodal Explanation}
This type of explanation utilizes multiple modalities to form a more comprehensive explanation for CMR processes.
The methods of multimodal explanation can be further classified into two categories: Independent Explanation and Joint Explanation.

\textbf{Independent Explanation}\quad These methods provide independent explanations of different modalities. According to the combination of modalities, the existing methods can be divided into three groups: (1) \textbf{Graph+Text}: These methods provide both graph explanations and textual explanations. (2) \textbf{Graph+Symbol}: These methods provide both graph explanations and symbol explanations. (3) \textbf{Visual+Text}: These methods provide both visual explanations and textual explanations.

\textbf{Joint Explanation}\quad These methods utilize multiple modalities to form a joint explanation for the CMR process. According to the combination of modalities, the existing methods belong to only one class: (1) \textbf{Visual+Text}: These methods combine visualization techniques and text to explain the CMR process more comprehensively.

\section{Visual Explanation}
Visual explanation methods adopt various visualization techniques to provide visual explanations for CMR processes.
These methods can promote the intuitive comprehension of the manner in which a CMR model understands the multimodal input and performs cross-modal interactions in the reasoning process.
According to the perspective of the model's reasoning process that these methods aim to visualize, we roughly categorize visual explanation methods into two subcategories: Single-modal Contribution and Cross-modal Relevance.
The methods of the first subcategory serve to visualize the contribution of the single-modal input to find key regions. Conversely, the methods of the second subcategory aim to visualize the relevance between multimodal inputs learned and further utilized by the reasoning model.

\subsection{Single-modal Contribution}

	\begin{figure}[ht]
		\begin{minipage}{0.59\linewidth}
			\includegraphics[width=1\linewidth]{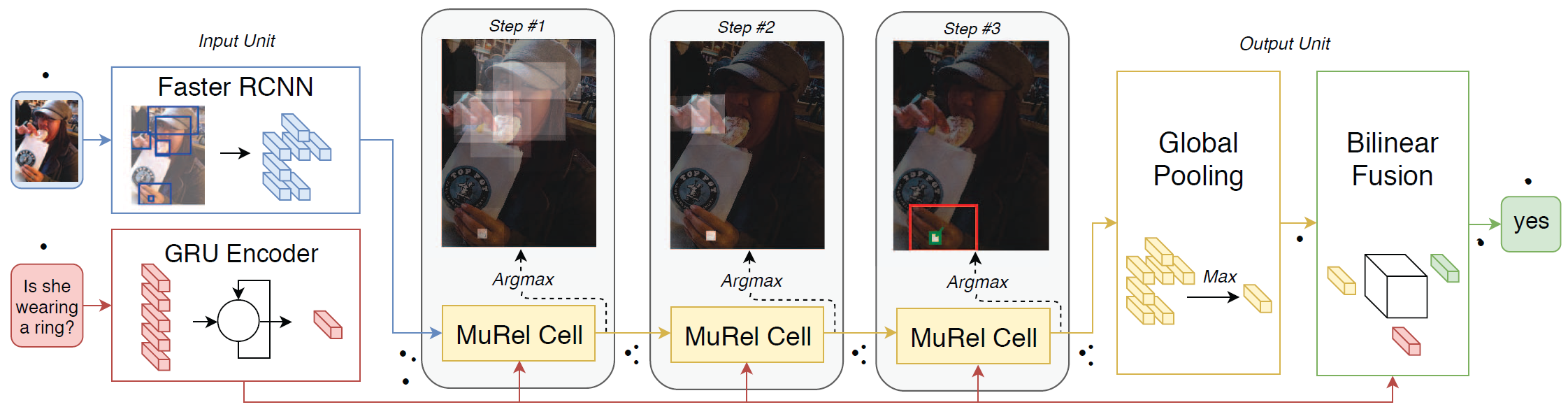}
			\caption{MuRel constructs the importance map of visual objects at each reasoning step by the proposed MuRel cell. The figure is from reference \cite{cadene2019murel}.}
			\label{fig:murele}
		\end{minipage}
		\hfill
		\begin{minipage}{0.39\linewidth}
			\includegraphics[width=1\linewidth]{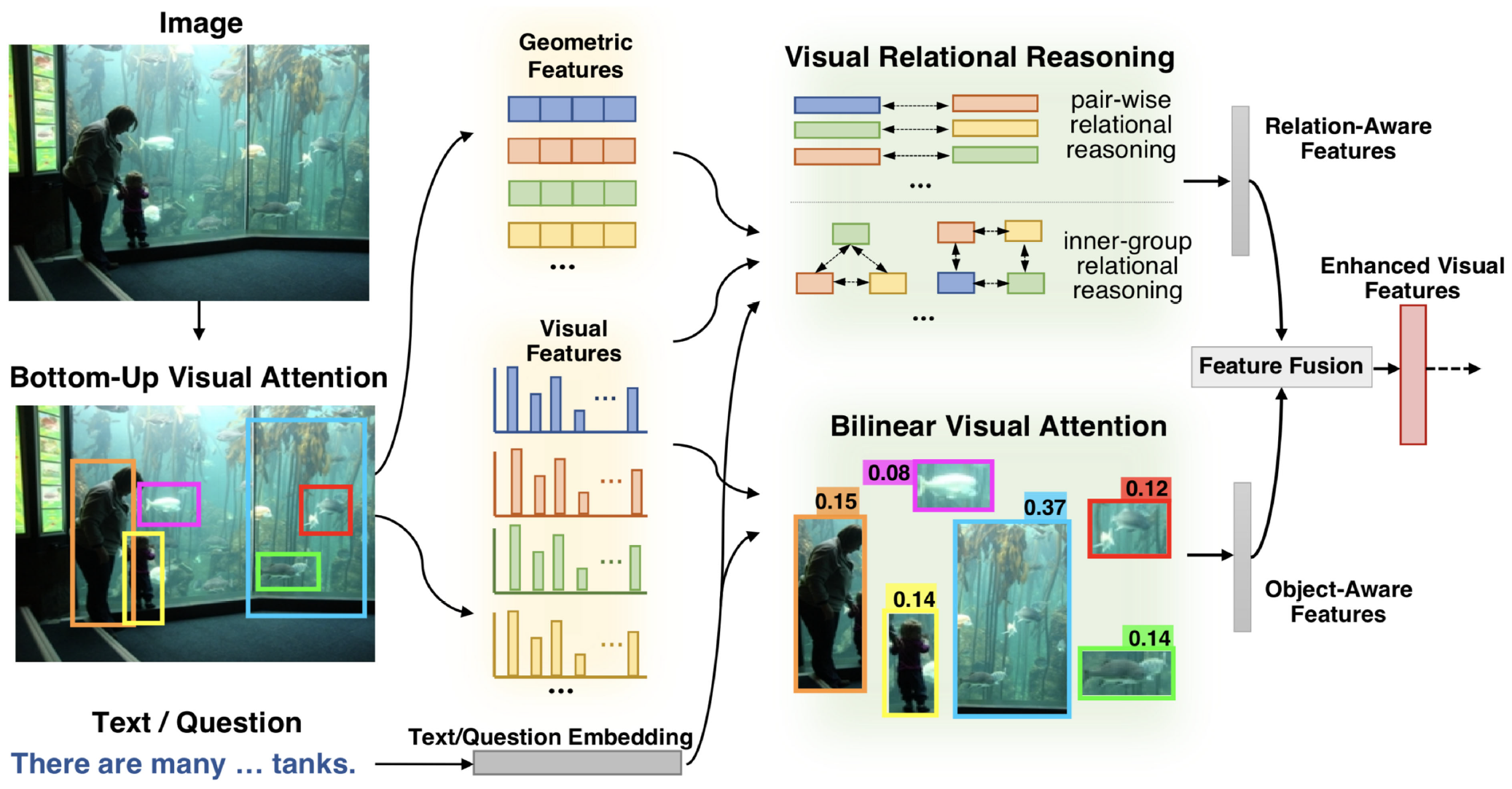}
			\caption{VRANet computes the visual attention for visual objects and fuses visual features using attention as the weights. The figure is from reference \cite{yu2020reasoning}.}
			\label{fig:36f}
		\end{minipage}
	\end{figure}
 
Cadene et al. \cite{cadene2019murel} introduce Multi-Modal Relationship Network (MuRel), an end-to-end learning approach for visual question answering (VQA), as shown in Fig. \ref{fig:murele}. The core of this model lies in the introduction of the MuRel cell, which utilizes vectors to represent the interactions between questions and image regions, modeling region relationships. MuRel adopts an iterative framework for VQA, which explicitly computes the importance of visual regions through rich vector representations of questions and visual data.

Yu et al. \cite{yu2020reasoning} propose a Visual Reasoning and Attention Network (VRANet) to capture rich visual semantics and enhance visual representations. As shown in Fig. \ref{fig:36f}, VRANet consists of two modules: the Visual Relationship Reasoning module enriches the representation of each object based on its relevance, and the Bilinear Visual Attention module identifies crucial targets based on textual content. VRANet can generate visualizations of the reasoning process, where the probability annotations in the image indicate the attention weights on the annotated objects.

Perez et al. \cite{perez2018film} propose  Feature-wise Linear Modulation (FiLM) to achieve visual reasoning. By utilizing FiLM layers to selectively and meaningfully manipulate the intermediate features of neural networks, RNNs can effectively use language to modulate CNNs for performing various multi-step reasoning tasks on images. The authors visualize the distribution of visual locations that the model uses for its globally max-pooled features to predict the final results.

When implementing visualization, starting from the critical local parts of the data is often advantageous. In the research of zero-shot sketch-based image retrieval (ZS-SBIR), Lin et al. \cite{lin2023zero} propose a visualization method that allows for retrieving corresponding images from user-provided sketches. They suggest that the cross-modal matching problem can be simplified into the concept of key local patch groups and propose the idea of utilizing cross-modal tokens for matching. This method uses tokens to achieve visualization of image data.

As the prevalence of ironic content continues to rise on the internet, the detection of multimodal irony has attracted the attention of many researchers. One such approach is unimodal and cross-modal graphs (InCrossMGs) \cite{liang2021multi}, which constructs two unimodal graphs and a cross-modal graph for each multi-modal example based on the hidden representations of text and image modalities. By providing attention visualization, InCrossMGs exhibits how to attend to incongruous regions between the image and the text.

Medical Visual Question Answering (Medical-VQA) primarily focuses on addressing questions related to medical radiological images. UnICLAM \cite{zhan2022uniclam}, an interpretable medical VQA model, adopts a unified dual-stream pre-training structure and a progressive soft parameter-sharing strategy. UnICLAM aligns image-text representation through contrastive representation learning and adversarial masking, thereby enabling more accurate and interpretable visual explanations.

Lyu et al. \cite{lyu2022dime} propose DIME, a method for interpreting multimodal models. The core idea of DIME is to provide more fine-grained interpretations by disentangling a multimodal model into unimodal contributions and multimodal interactions. DIME gains unimodal contributions from information gained by only looking at one of the modalities without interacting with any other modalities and generates visualization both on text and vision. The visualization shows how unimodal contributions influence the decision of the models.

MULTIVIZ \cite{liang2023multiviz} is a method for analyzing the behavior of multi-modal models, providing an interactive visualization API across multi-modal datasets and models. MULTIVIZ understands the contributions of textual and visual modalities toward modeling and prediction and generates a variety of visualizations to aid in understanding multimodal models, such as Attention Maps, Saliency Maps, etc.

	\begin{figure}[ht]
		\begin{minipage}{0.44\linewidth}
			\includegraphics[width=1\linewidth]{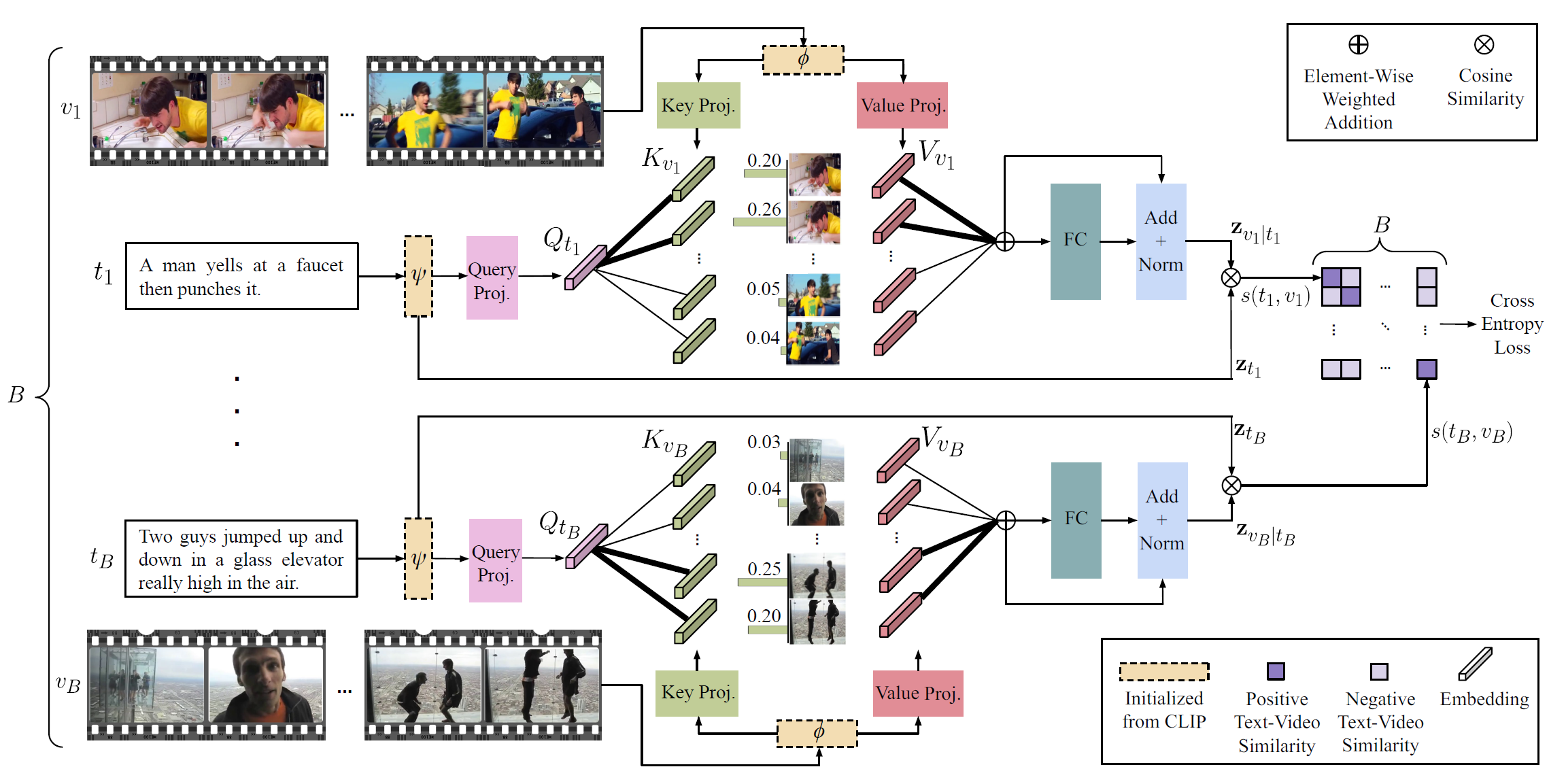}
			\caption{For text-video retrieval task, X-Pool visualizes the cross-modal attention of text query to each frame of the video. The figure is from reference \cite{gorti2022x}.}
			\label{fig:44o}
		\end{minipage}
		\hfill
		\begin{minipage}{0.54\linewidth}
			\includegraphics[width=1\linewidth]{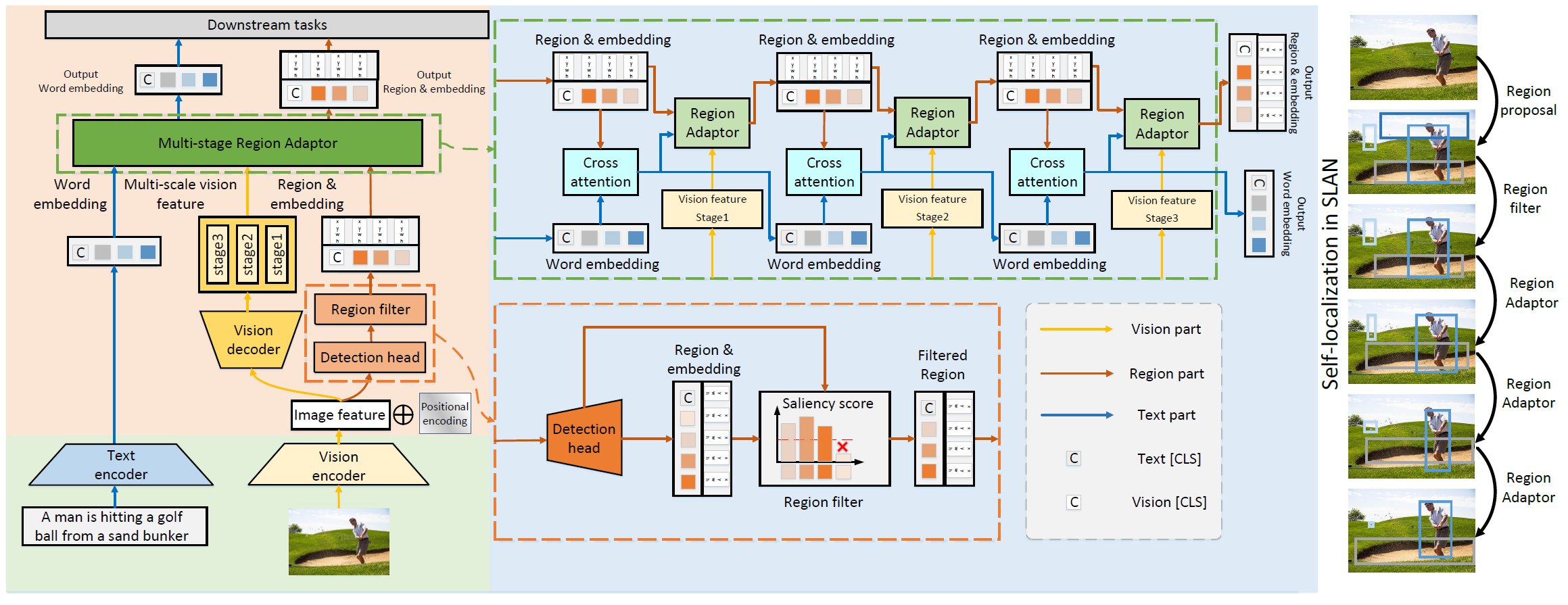}
			\caption{SLAN utilizes a self-locator to predict region saliency with a region filter and progressively grounds salient regions. The figure is from reference \cite{zhai2022slan}.}
                \label{fig:45o}
		\end{minipage}
	\end{figure}

\subsection{Cross-modal Relevance}
Gorti et al. \cite{gorti2022x} study text-video retrieval and propose a cross-modal attention model named X-Pool. Specifically, as illustrated in Figure \ref{fig:44o}, the X-Pool model extracts video features and encodes textual queries. Afterward, it employs an attention mechanism to weigh and integrate the semantic information between video frames and text queries. X-Pool explains the correlation between text and video, enabling inference and matching, and allowing for the generation of visual explanations.

When considering the interaction between video and text, Jin et al. \cite{ jin2023video} evaluate the potential correspondence between video frames and text words and propose a hierarchical Banzhaf interaction (HBI). The core idea of the method is to treat video and text data as players in a game and simulate their interactions to learn their communication and relationship. Specifically, they propose a hierarchical interaction network to capture the correlation between video and text through multi-level interactions. 

To bridge the semantic gap between images and texts, Ji et al. \cite{ DBLP:conf/ijcai/JiCW21} attempt to decompose the image-text matching process into a multi-step cross-modal reasoning process, and propose a Stepwise Hierarchical Alignment Network (SHAN). SHAN first achieves local-to-local alignment at the fragment level, then performs global-to-local alignment at the context level, and finally achieves global-to-global alignment. The alignment process essentially corresponds image with text, providing a visual explanation for the direct relationship between images and text.

Ren et al. \cite{ren2021learning}, who also focus on the relationship between words and objects in the image, introduce a metric called Intra-modal Self-attention Distance (ISD). ISD measures the semantic distance between linguistic relationships and visual relationships. Building upon the optimization of ISD, they further propose an Inter-modal Alignment through Intra-modal Self-attention (IAIS) approach. IAIS aims to calibrate the intra-modal self-attention of both modalities by aligning them with each other.

Based on the requirements of the vision-language task, Zhai et al. \cite{zhai2022slan} propose a self-localization assisted network (SLAN) for cross-modal understanding tasks, such as image-text retrieval and phrase grounding, and achieve visual explanation. The overall framework of SLAN is depicted in Figure \ref{fig:45o}. SLAN enables visual interpretation through its framework, addressing the demands of vision-language tasks.

Yang et al. \cite{yang2020robust} propose a model called t-RNetAttn for handling spatial relations in the context of textual information. The model employs a relational network that computes representations for each position in the environment, implicitly enabling factorization through interactions with neighboring entities. By utilizing vectorized representations derived from the textual input and attention data, it visualizes value maps and relation graphs.

	\begin{figure}[ht]
		\begin{minipage}{0.49\linewidth}
			\includegraphics[width=1\linewidth]{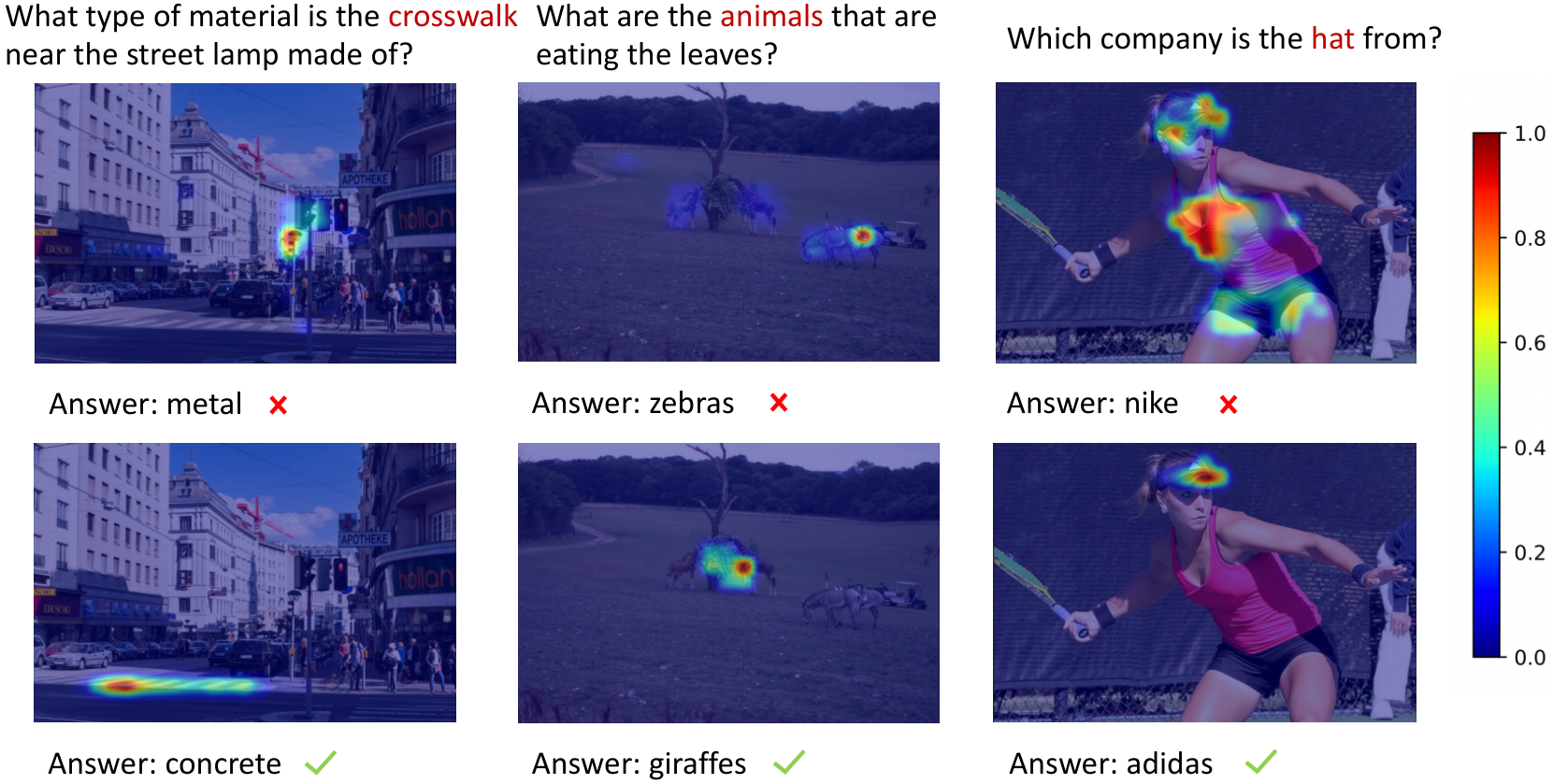}
                \caption{For the VQA task, CRRN visualizes the visual attention to the keyword in the input question, investigating the model's capacity to focus on the relevant regions. The figure is from reference  \cite{chen2021cross}.}
                \label{fig:41e}
		\end{minipage}
		\hfill
		\begin{minipage}{0.49\linewidth}
			\includegraphics[width=1\linewidth]{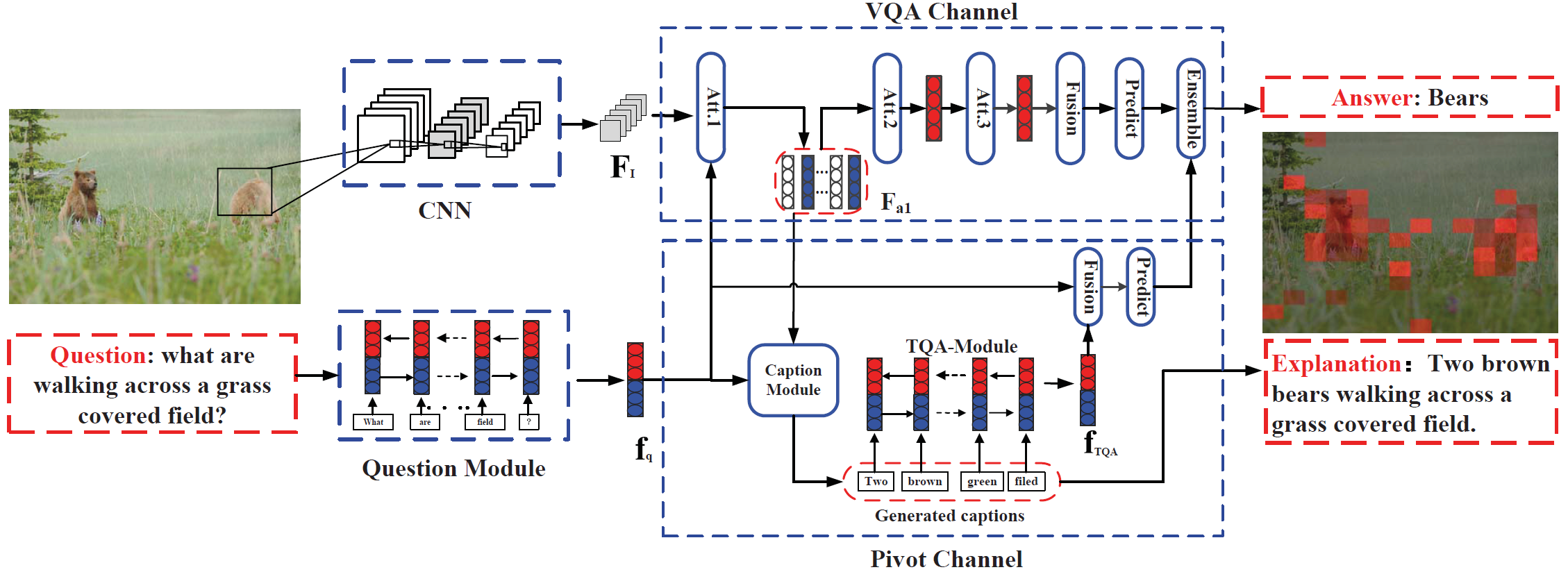}
                \caption{NPN utilizes a caption module to describe the image based on the question for visual question answering. The feature of the caption explanation is integrated to predict the answer. The figure is from reference \cite{zhou2017more}.}
                \label{fig:48NPN}
		\end{minipage}
	\end{figure}
 
To address the VQA task, a cross-modal understanding of both images and questions is necessary. Chen et al. \cite{chen2021cross} propose a Cross-modal Relational Reasoning Network (CRRN). CRRN employs visual grounding to comprehend the content within the image, facilitating interaction between image regions and words. Through relational reasoning (see Figure \ref{fig:41e}), it infers semantic information that represents the association between the image and the question.

Xue et al. \cite{xue2022mmt} propose a multi-modal memory transformer (MMT) framework for Image-guided Story Ending Generation (IgSEG). MMT utilizes a memory-augmented cross-modal attention network to capture relations between text and image. Additionally, MMT provides visual explanations of the cross-modal attention network. By calculating the attention weight of each word towards all image regions and vice versa, they visualize the important words and regions in multimodal input. These heatmaps aid in understanding the reasoning process of MMT.

DIME \cite{lyu2022dime} gains multimodal interactions from cross-referencing inputs from multiple modalities. Visually, the weights calculated by DIME are used to generate a human-interpretable visualization: for images, each feature is typically a part of the image, so the critical parts with high weights can be highlighted. For text, each feature is typically a word, so the explanation can be summarized as a histogram of the weights of each word.

MULTIVIZ \cite{liang2023multiviz} divides the interpretability problem into four stages:  Single-modal Importance, Cross-modal interaction, Multi-modal Representation, and Multi-modal Prediction. The stage of Cross-modal interaction aims to gain Cross-modal interactions to describe various ways in which atoms from different modalities can relate with each other and the types of new information possibly discovered as a result of these relationships. MULTIVIZ visualizes multimodal representations through local and global analysis.

\begin{table}[ht]
\centering
\caption{Categorization of visual explanation methods.}
\begin{tabular}{|c|c|c|}
\hline
\textbf{Category}                                                                    & \textbf{Subcategory}                          & \textbf{Method} \\ \hline
\multirow{9}{*}{Single-modal Contribution} & \multirow{3}{*}{Visual Object Contribution}   & MuRel \cite{cadene2019murel}      \\ \cline{3-3} 
                                                                                     &                                               & VRANet \cite{yu2020reasoning}           \\ \cline{3-3} 
                                                                                     &                                               & VizWiz-VQA \cite{chen2022grounding}          \\ \cline{2-3} 
                                                                                     & \multirow{2}{*}{Visual Grid Contribution}    & FiLM \cite{perez2018film}            \\ \cline{3-3} 
                                                                                     &                                               & ZS-SBIR \cite{lin2023zero}         \\ \cline{2-3} 
                                                                                     & \multirow{4}{*}{Region and Word Contribution} & InCrossMGs \cite{liang2021multi}      \\ \cline{3-3} 
                                                                                     &                                               & UnICLAM \cite{zhan2022uniclam}         \\ \cline{3-3} 
                                                                                     &                                               & DIME \cite{lyu2022dime}            \\ \cline{3-3} 
                                                                                     &                                               & MultiViz \cite{liang2023multiviz}        \\ \hline
\multirow{10}{*}{Cross-modal Relevance}    & \multirow{2}{*}{Text-and-Frame Relevance}                & X-Pool \cite{gorti2022x}          \\ \cline{3-3} 
                                                                                     &                                               & HBI \cite{jin2023video}             \\ \cline{2-3} 
                                                                                     & \multirow{3}{*}{Phrase-and-Object Relevance}             & SHAN \cite{DBLP:conf/ijcai/JiCW21}            \\ \cline{3-3} 
                                                                                     &                                               & IAIS \cite{ren2021learning}            \\ \cline{3-3} 
                                                                                     &                                               & SLAN \cite{zhai2022slan}            \\ \cline{2-3} 
                                                                                     & \multirow{5}{*}{Phrase-and-Grid Relevance}               & t-RNetAttn \cite{yang2020robust}      \\ \cline{3-3} 
                                                                                     &                                               & CRRN \cite{chen2021cross}            \\ \cline{3-3} 
                                                                                     &                                               & MMT \cite{xue2022mmt}             \\ \cline{3-3} 
                                                                                     &                                               & DIME \cite{lyu2022dime}            \\ \cline{3-3} 
                                                                                     &                                               & MultiViz \cite{liang2023multiviz}        \\ \hline
\end{tabular}
\label{tab:vis}
\end{table}

\subsection{Discussion}
The objective of visual explanation is to provide intuitive visualization results that can reveal and elucidate certain perspectives of the CMR process.
These explanations alleviate the black-box nature of the neural network model and offer an approach for users to understand and inspect the reasoning processes of the model.
We summarize the abovementioned methods in Table \ref{tab:vis}. 
Methods of single-modal contribution mostly visualize the contribution of visual objects, visual grids, or visual regions and words.
Methods of cross-modal relevance typically visualize the relevance between texts and frames, phrases and objects, or phrases and grids.
Moreover, some common similarities shared by visual explanation methods can be listed as follows:

\begin{itemize}
    \item \textbf{Quantification of Visualized Regions}\quad Irrespective of the specific kind of regions (e.g., image grids, visual objects, and textual words) under consideration, visual explanation methods need to quantify the visualized regions. The quantification values include the abovementioned single-modal contribution and cross-modal relevance. Subsequently, the quantification values are visualized by various visualization techniques.
    
    \item \textbf{Heatmap}\quad A large amount of visual explanation methods utilize the heatmap to illustrate the magnitude of single-modal contribution or cross-modal relevance, such as \cite{perez2018film, cadene2019murel, yang2020robust, chen2021cross, liang2021multi, xue2022mmt, zhai2022slan}. Heatmaps employ color gradations, ranging from cool to warm tones, to represent varying magnitudes, which provides an intuitive visual explanation.

    \item \textbf{Attention}\quad Methods for cross-modal relevance mostly employ attention mechanisms to model the relations across modalities. The obtained attention values serve as quantifiable indicators of cross-modal relevance and are subsequently subjected to visualization techniques.

\end{itemize}
Visual explanation methods offer intuitive explications for CMR processes, which users can quickly comprehend. However, existing methods of visual explanation may suffer from the following limitations:
\begin{itemize}
    \item \textbf{Inaccuracy}\quad It is well-known that accurately computing the contribution of input regions remains a challenging problem \cite{selvaraju2017grad, srinivas2019full, jiang2021layercam}. Moreover, since recent methods with attention mechanisms mostly adopt multiple attention layers, appropriately visualizing the multi-layer cross-modal attention is complicated and tricky. Consequently, the provided visual explanation may not accurately interpret the intricacies of CMR processes.

    \item \textbf{Limited Interpretability}\quad Existing visual explanation methods merely provide insights into single-modal contribution and cross-modal relevance, which are insufficient to explain the reasoning process of CMR models. For example, MULTIVIZ \cite{liang2023multiviz} can identify pivotal regions and their relationships in multimodal input. However, it remains limited in its ability to articulate the precise utilization of these identified regions and relationships in deriving the ultimate reasoning outcomes.

\end{itemize}

\section{Textual Explanation}

Textual explanation methods provide explanations in the form of natural language for CMR processes, which can be easily understood by users. It aids human understanding of the model’s reasoning process and results. 
While reviewing textual explanation methods, we categorize various methods according to the content that their generated explanations aim to convey. 
On the one hand, some methods attempt to explain the facts for the input by text, such as describing the visual input and retrieving basic knowledge for the reasoning. 
On the other hand, some methods attach greater importance to the reasoning process itself, explaining the reasoning process in detail. 
Hence, we can roughly categorize textual explanation methods into two groups: Facts for Input and Description for Reasoning Process.

\subsection{Facts for Input}
Zhou et al. \cite{zhou2017more} propose a model called Neural Pivot Network (NPN), consisting of several components (see Figure \ref{fig:48NPN}) and generating explanations in a multi-task learning architecture. In the pivot channel, a caption module is utilized to generate an internal caption. They use both image-caption and image-question-answer pairs to train NPN. The model provides a question-based sentence to explain its prediction.

Li et al. \cite{li2018tell} propose a Tell and Answer (TAA) model for VQA, consisting of three modules: word prediction, sentence generation, and answer reasoning. In the word prediction module, the image is fed into a pre-trained visual detector to extract word-level explanations. In the sentence generation module, the image is inputted into a pre-trained caption model to generate sentence-level explanations. Finally, a reasoning module finishes answer prediction through these explanations.

Li et al. \cite{li2018vqa} propose a VQA-E (VQA with Explanation) task, where the model is required to generate textual explanations based on the predicted answers. Initially, the image is represented by a pre-trained CNN, while the question is encoded. Subsequently, the image and question features are fed into an attention module to obtain image features related to the question. Finally, the answers and textual explanations are generated simultaneously using the question and image features.

In comparison to the difficulty of multimodal feature fusion, Li et al. \cite{li2019visual} attempt to address VQA from the perspective of machine reading comprehension. They propose using natural language to unify all input information, thereby transforming VQA problems into machine reading comprehension problems. The proposed approach transforms visual information into textual descriptions. By conducting multimodal fusion in the textual domain, valuable semantic information relevant to visual question answering is preserved.

Tseng et al. \cite{tseng2022relation} propose an interpretable model for the VQA task, which is based on input facts. This model is divided into four modules: a feature extraction module, a relation encoder, an explanation generator, and a caption-attended predictor. In the explanation generator, an LSTM layer serves as the core to describe the relationships among the extracted visual features from the images and generates question-related facts in textual form as explanations.

Guo et al. \cite{guo2023images} propose Img2LLM to enable Large Language Models (LLMs) to perform zero-shot VQA tasks without end-to-end training. Img2LLM utilizes an image-question matching module and a caption model to transform the image into a caption prompt. Then, Img2LLM conducts answer extraction and synthetic question generation based on the caption prompt to generate an exemplar prompt. The textual prompts with the question are input into an LLM to perform partial VQA tasks. 

In order to utilize external knowledge for fact generation, Wang et al. \cite{wang2017fvqa} propose  Fact-based VQA (FVQA) to answer visual questions that require external information to answer. Firstly, their method extracts visual attributes from images and links them to corresponding semantic entities. Then, using an LSTM model, they map the input questions to a specific query type. Finally, the facts are matched with keywords in the questions to select the best-matching fact and obtain the answer. This method generates facts to serve as a textual explanation.

\begin{figure}[ht]
 \begin{minipage}{0.54\linewidth}
    \includegraphics[width=1\linewidth]{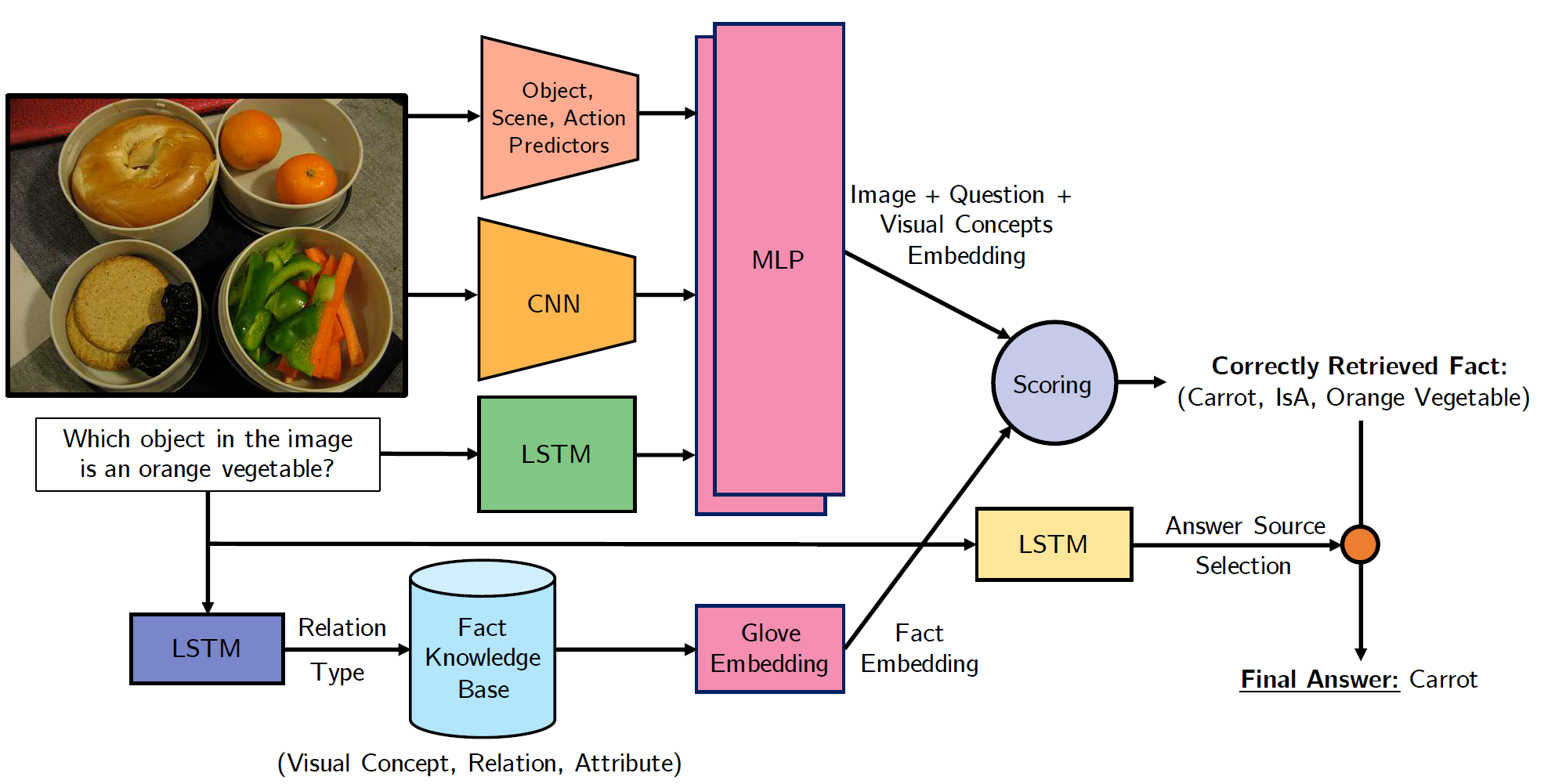}
    \caption{LKBR uses an LSTM to predict the fact relation type from the question and retrieves structured facts from a knowledge base. The figure is from reference \cite{narasimhan2018straight}.}
    \label{fig:50s}
\end{minipage}
\hfill
\begin{minipage}{0.44\linewidth}
    \includegraphics[width=0.8\linewidth]{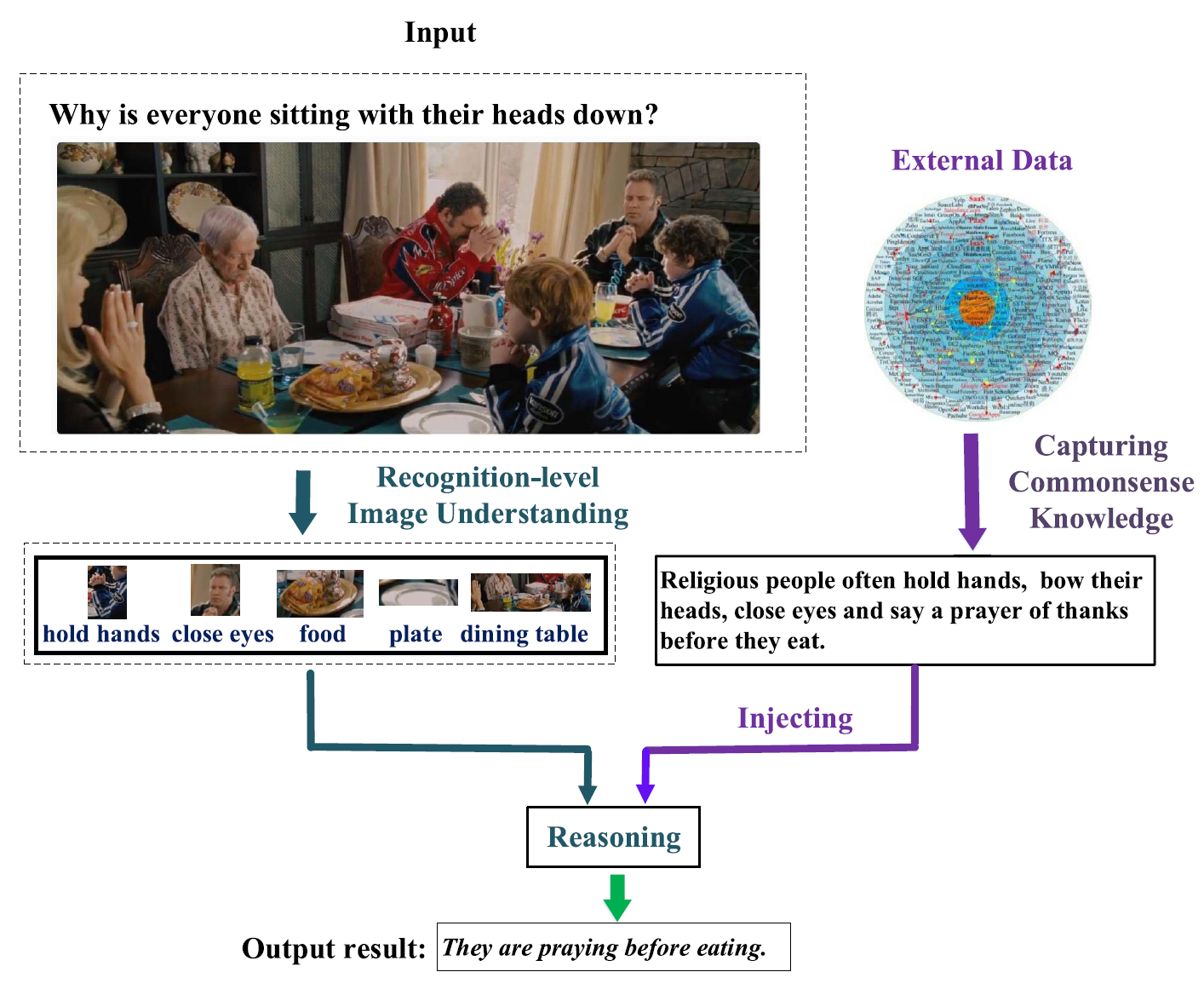}
    \caption{CKRM retrieves commonsense knowledge text to assist the reasoning for visual questions. The figure is from reference \cite{wen2020multi}.}
    \label{fig:53e}
\end{minipage}
\end{figure}

To select facts that are more closely aligned with the image-question pairs, a learning-based retrieval approach has been proposed \cite{narasimhan2018straight}. This method is illustrated in Fig. \ref{fig:50s}. 
In the model, a multi-layer perceptron (MLP) is used to fuse multimodal input features and obtain an image-question embedding. The retrieved facts are ranked by computing the dot product between the image-question embeddings and the fact embeddings. The retrieved facts are utilized to predict the final result and can serve as a textual explanation for reasoning.

In Fig. \ref{fig:53e}, Wen and Peng \cite{wen2020multi} propose the Knowledge-based Reasoning Model (CKRM) to acquire external knowledge for addressing more challenging commonsense problems. This model uses a Multi-level Knowledge Transfer Network to gather commonsense knowledge from source tasks. Then, a knowledge-based reasoning approach further utilizes the transferred knowledge to support visual commonsense reasoning.

Zhang et al. \cite{zhang2021explicit} incorporate syntactic information into visual reasoning and language semantic understanding and propose an explicit cross-modal representation learning network. This approach consists of two key modules: the syntactic sensitive text understanding module models the semantic structure of sentences and extracts textual representations; the syntactic guided cross-modal reasoning progressively focuses on relevant objects using multiple neural modules based on attention mechanisms.

\subsection{Description for Reasoning Process}
%
%

Marasovic et al. \cite{marasovic2020natural} propose an integrated model called Rationale Transformer, which learns to generate free-text reasoning by combining a pre-trained language model with object recognition, a basic visual semantic framework, and a visual commonsense graph. In the model, they first employ an object detector to predict the objects present in the image. Then, they use a model for grounded situation recognition. Finally, they leverage the VisualComet model, which is based on GPT-2 and capable of generating commonsense inferences.

Recognizing that answers and rationales can mutually influence each other, Dua et al. \cite{dua2021beyond} propose an attention-based end-to-end encoder-decoder architecture. This model consists of four complete LSTM-based sub-modules. The first two sub-modules are used to generate answers and produce rationales along with their representations. In the latter two sub-modules, leveraging the features from the previous outputs, more refined answers and corresponding reasoning rules are generated.

Lu et al. \cite{lu2022learn} introduce a new benchmark that consists of a large number of multiple-choice science questions on various scientific topics, along with answers and corresponding explanations. Based on ScienceQA, they design language models to learn to generate lectures and explanations as a chain of thought (CoT) to simulate the multidimensional reasoning process of ScienceQA. By adopting this thinking process, they adapt two language models, UnifiedQA with the CoT, and GPT-3 via CoT prompting, to generate textual content for the intermediate reasoning process.

\begin{figure}[ht]
 \begin{minipage}{0.54\linewidth}
    \includegraphics[width=1\linewidth]{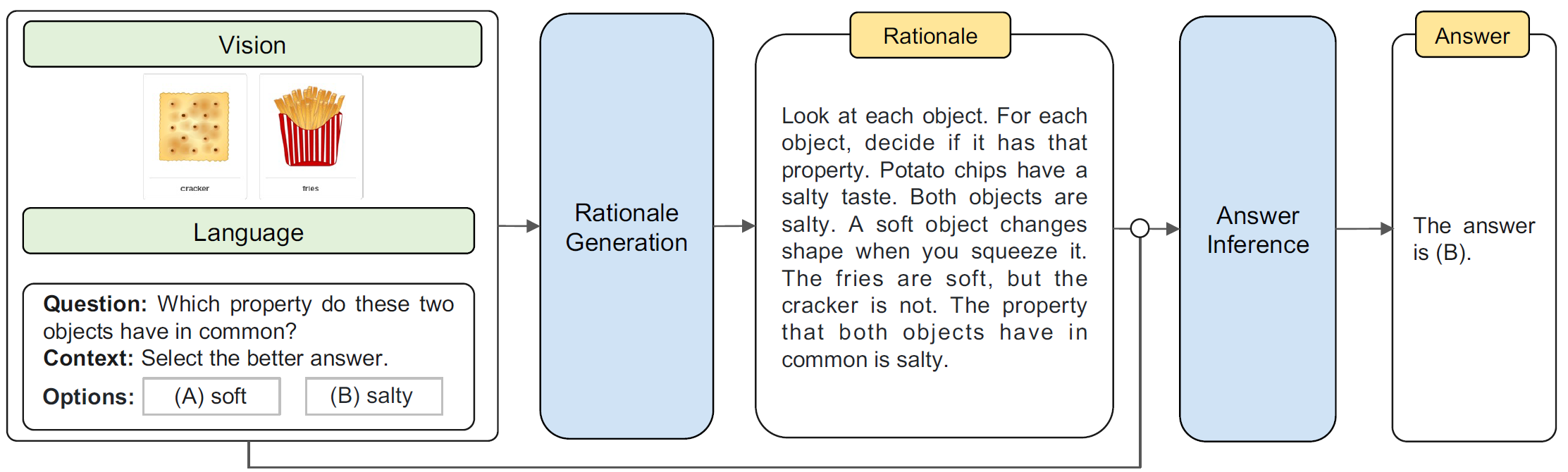}
    \caption{Multimodal-CoT adopts a multimodal Transformer to generate the step-by-step rationale and then predict the answer. The figure is from reference \cite{zhang2023multimodal}.}
    \label{fig:24o}
\end{minipage}
\hfill
\begin{minipage}{0.44\linewidth}
    \includegraphics[width=1\linewidth]{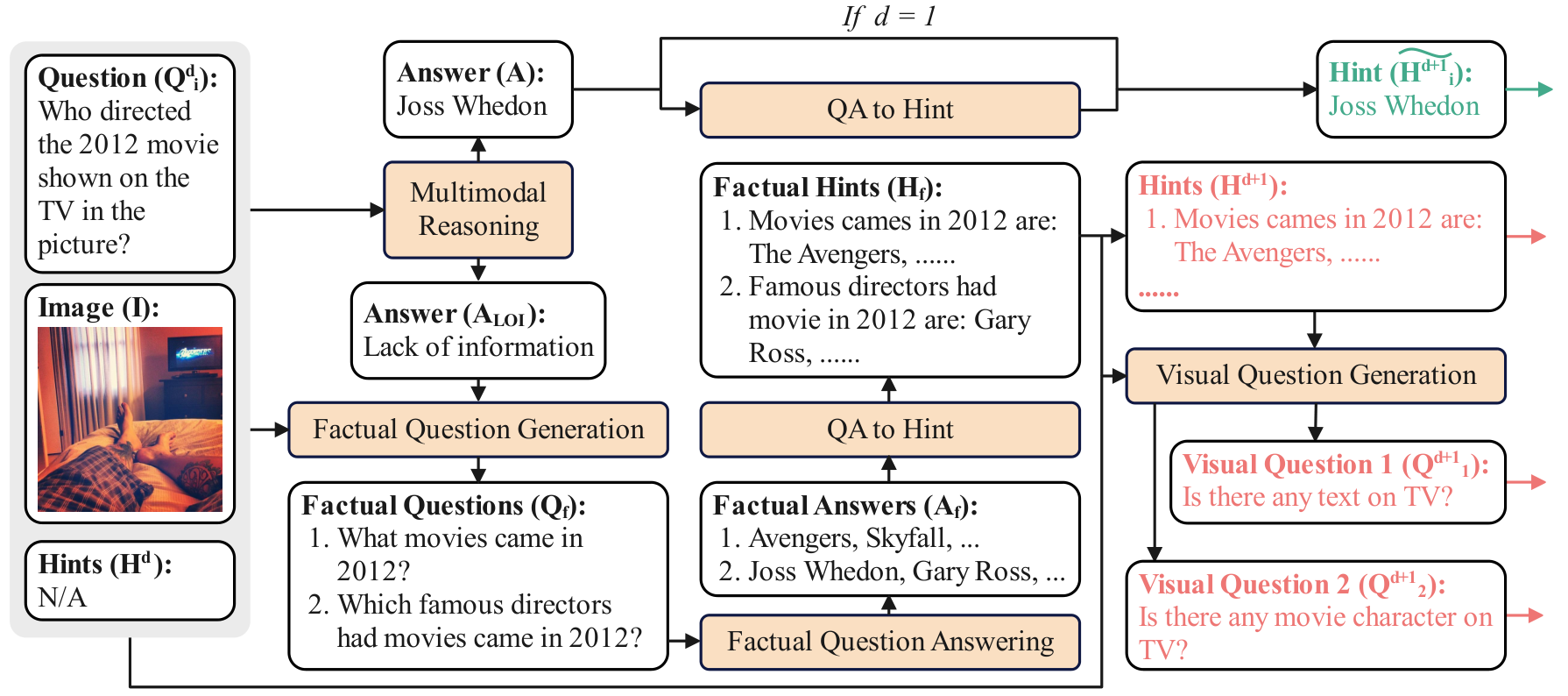}
    \caption{SOCRATIC generates sub-problems related to the original problem and recursively backtracks and answers the sub-problems until reaches the original problem. The figure is from reference \cite{qi2023art}.}
    \label{fig:59o}
\end{minipage}
\end{figure}

Inspired by LLMs, CoT has the effect of inducing intermediate reasoning chains in the field of multimodal reasoning \cite{zhang2023multimodal}. Zhang et al. propose Multimodal-CoT, which combines visual and textual modalities for reasoning while separately generating answers and rationales. As shown in Fig. \ref{fig:24o}, their model consists of two main stages. Especially, it is worth noting that the model generates synthetic question-answer pairs as in context exemplars from the current image in the rationale generation stage.

Previous studies have explored the application of CoT reasoning in complex multimodal scenarios. However, this approach is limited by the costly process of collecting high-quality CoT principles and the issue of annotation inaccuracies. To address these challenges, Wang et al. \cite{wang2023t} propose a novel method called T-SciQ. This approach generates high-quality CoT rationales as teaching signals and performs CoT reasoning in complex modalities.

Qi et al. \cite{qi2023art} propose an algorithm called SOCRATIC to simulate the human thinking process, both top-down and bottom-up, as shown in Fig. \ref{fig:59o}. The Socratic questioning method is essentially a recursive thinking process that involves both top-down exploration and bottom-up backtracking. The core component of this algorithm is a Self-Questioning module that utilizes LLMs to actively generate and answer a series of relevant questions. Through this approach, the intermediate reasoning steps can be represented in textual form.

\begin{table}[ht]
\centering
\caption{Categorization of textual explanation methods.}
\begin{tabular}{|c|c|c|}
\hline
\textbf{Category}                                                                             & \textbf{Type of Texts}            & \textbf{Method} \\ \hline
\multirow{10}{*}{Facts for Input}                                                             & \multirow{6}{*}{Generated Facts} & NPN \cite{zhou2017more}         \\ \cline{3-3} 
                                                                                              &                                 & TAA \cite{li2018tell}           \\ \cline{3-3} 
                                                                                              &                                 & VQA-E \cite{li2018vqa}             \\ \cline{3-3} 
                                                                                              &                                 & VQARC \cite{li2019visual}           \\ \cline{3-3} 
                                                                                              &                                 & RAIC \cite{tseng2022relation}            \\ \cline{3-3} 
                                                                                              &                                 & Img2LLM \cite{guo2023images}             \\ \cline{2-3} 
                                                                                              & \multirow{4}{*}{Retrieved Facts} & FVQA \cite{wang2017fvqa}           \\ \cline{3-3} 
                                                                                              &                                 & LKBR \cite{narasimhan2018straight}           \\ \cline{3-3} 
                                                                                              &                                 & CKRM \cite{wen2020multi}            \\ \cline{3-3} 
                                                                                              &                                 & ECMRL \cite{zhang2021explicit}            \\ \hline
\multirow{6}{*}{Description for Reasoning Process} & \multirow{2}{*}{Concise Rationale} & Rationale Transformer \cite{marasovic2020natural} \\ \cline{3-3} 
                                                                                              &                                    & ViQAR \cite{dua2021beyond} \\ \cline{2-3} 
                                                                                              & \multirow{3}{*}{Chain-of-Thought}  & ScienceQA \cite{lu2022learn}        \\ \cline{3-3} 
                                                                                              &                                    & Multimodal-CoT \cite{zhang2023multimodal}        \\ \cline{3-3} 
                                                                                              &                                    & T-SciQ \cite{wang2023t}                \\ \cline{2-3} 
                                                                                              & Self-Questioning                   & SOCRATIC \cite{qi2023art}              \\ \hline
\end{tabular}
\label{tab:tex}
\end{table}

\subsection{Discussion}
The core target of textual explanations in CMR is to provide users with accurate, understandable, and trustworthy explanations of the CMR process. This promotes the interpretability of the model and builds user trust. Different models may have different emphases and implementations for CMR. 
We categorize the textual explanation methods mentioned above according to the type of textual explanations in Table \ref{tab:tex}.
Moreover, some similarities of textual explanation methods can be concluded as follows:

\begin{itemize}
    \item \textbf{Fusing Textual Explanations}\quad Though the generated or retrieved textual explanations can independently interpret the reasoning process, most methods fuse the feature of textual explanation into the reasoning. 
    The fusion of textual explanations can improve the consistency between the explanation and the real reasoning process of models as well as the effectiveness of reasoning.

    \item \textbf{Legibility}\quad Legibility of the explanations is a shared strength of textual explanation methods.
    Notably, recent methods have shown promising abilities in describing the reasoning process in a comprehensive manner.
    Since the provided explanations are readily comprehensible to users, these methods exhibit considerable potential for constructing CMR models with superior interpretability.


\end{itemize}

Textual explanations can be easily understood by users, which suggests the feasibility of textual explanation methods in real-world applications. However, existing methods may suffer from the following limitations:
\begin{itemize}

    \item \textbf{Limited Cross-modal Interpretability}\quad The interpretability of textual explanation can potentially be constrained when explaining non-textual rationales. 
    For example, the text is not always capable or appropriate for describing detailed visual concepts.
    Meanwhile, humans usually incorporate concepts from multiple modalities in CMR, which suggests that text may be insufficient in fully explaining the process of CMR.
    
    \item \textbf{Ignorance of Explanation Evaluation}\quad Though numerous metrics are prevalent in the realm of text generation (e.g., BLEU \cite{papineni2002bleu}, METEOR \cite{banerjee2005meteor}, CIDEr \cite{vedantam2015cider}, and ROUGE \cite{lin2004rouge}), existing methods of textual explanation mostly ignore the evaluation of explanations.
    Consequently, the reliability and accuracy of the provided textual explanations may be doubtful.
    Conversely, considering the evaluation of textual explanations may serve as a prospective avenue for future research.

\end{itemize}


\section{Graph Explanation}
Graph explanation methods elucidate the CMR process by constructing graphs to represent the entities and their interrelations within the multimodal input.
According to whether the graph(s) is constructed using the input information from a single modality or multiple modalities, graph explanations can be divided into two distinct types: Single-modal Graph and Multi-modal Graph.
%

\subsection{Single-modal Graph}
Norcliffe-Brown et al. \cite{norcliffe2018learning} propose a graph-based approach for VQA. They integrate a graph learner module that acquires a question-specific graph representation of the input image with the contemporary notion of graph convolutions, with the objective of acquiring image representations that effectively capture question-specific interactions. This approach enables the acquisition of question-specific object representations, which are influenced by relevant neighbors without requiring any manual description of the graph structure.

Li et al. \cite{li2019relation} focus on comprehending the visual scene depicted in images. They delineate two types of object relations, namely explicit relations which convey information pertaining to the geometric positioning and semantic interaction between objects, and implicit relations which capture the concealed dynamics existing between image regions. To acquire question-sensitive relation representations, they devise a Relation-aware Graph Attention Network (ReGAT) that employs a graph attention mechanism to model multi-type inter-object relations.

\begin{figure}[ht]
 \begin{minipage}{0.59\linewidth}
    \includegraphics[width=1\linewidth]{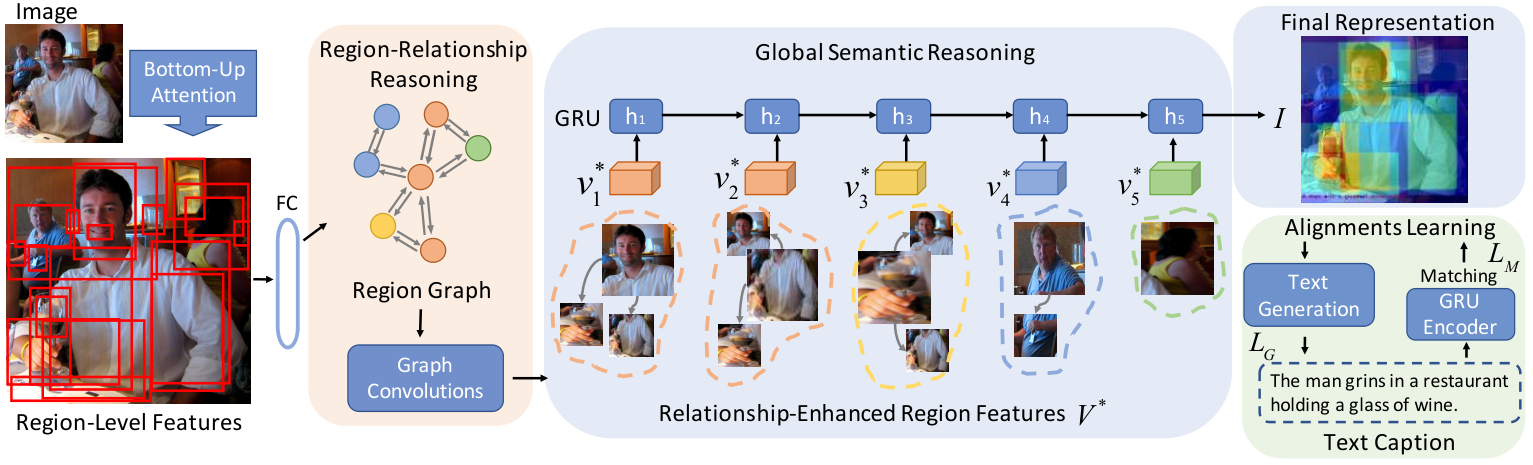}
    \caption{For image-text matching task, VSRN constructs a region graph for the image and computes the matching score based on the graph feature of the image and the GRU feature of the text. The figure is from reference \cite{li2019visual2}.}
    \label{fig:vsrn}
\end{minipage}
\hfill
\begin{minipage}{0.39\linewidth}
    \centering
    \includegraphics[width=0.6\linewidth]{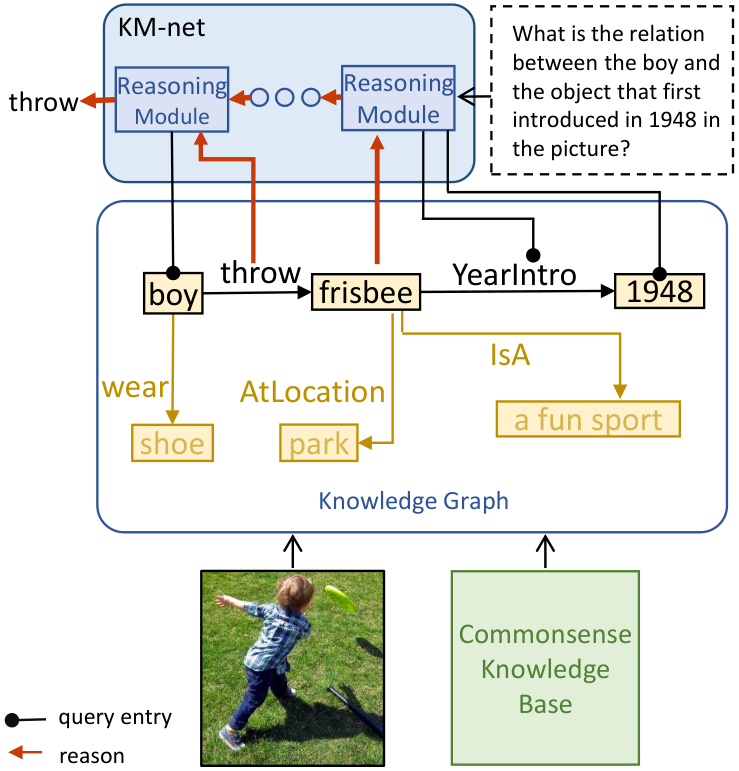}
    \caption{KM-net constructs a scene graph for the input image, which is further augmented using a knowledge base. The graph path (i.e., $boy\rightarrow frisbee\rightarrow 1948$) can explain the reasoning process. The figure is from reference \cite{cao2019explainable}.}
    \label{fig:kmnet}
\end{minipage}
\end{figure}

As shown in Fig. \ref{fig:vsrn}, Li et al. \cite{li2019visual2} establish region-region connections and utilize Graph Convolutional Networks to carry out reasoning on the region graph, thereby generating feature representations with enhanced semantic relationships. Subsequently, they introduce the gate and memory mechanism to conduct global semantic reasoning on these relationship-enhanced features, capturing the discriminative information and progressively synthesizing a representation of the entire scene.

Nguyen et al. \cite{nguyen2022coarse} establish a scene graph for the input image by Faster R-CNN, including Region Of Interest (ROI) features of objects, object attributes, and relations between objects. The obtained features are fed into the coarse-to-fine reasoning module to jointly learn their features and predicates. The predicates here are used to describe the content regarding objects, relationships, or image/question attributes. The coarse-grained and fine-grained features are integrated to predict the final answer. 

Cao et al. \cite{cao2019explainable} identify a deficiency in the interpretability aspect of the previous VQA benchmarks, where only a single accuracy metric is being utilized. In response, they propose the High-Order Visual Question Reasoning (HVQR) benchmark, aimed at assessing the VQA model’s ability to tackle explainable and high-order visual problems. 
Leveraging the HVQR benchmark, they propose a Knowledge Modularization Network (KM-net), as shown in Fig. \ref{fig:kmnet}. KM-net constructs a scene graph for the input image, which is further augmented using a knowledge base.

By leveraging both visual input and external knowledge bases, it is possible to extract high-level semantic information and relationships.  Building upon this conceptual framework, Zhang et al. \cite{zhang2022query} extracted both scene graphs and knowledge graphs. To enhance the generated queries with knowledge derived from the visual input and external knowledge base, they utilized scene graphs and knowledge graphs as inputs for the reasoning network.

Vatashsky and Ullman  \cite{vatashsky2020vqa}  propose a model without question-answer training consisting of a question-to-graph mapper and an answering procedure. On the one hand, the question-to-graph mapper utilizes a sequence-to-sequence LSTM model to map the questions into a graph representation. On the other hand, the answering procedure searches for valid assignments within the image based on the graph, generating answers in this manner. Their approach employs a unimodal graph, where the graph components include object classes, properties, and relationships.

\begin{figure}[ht]
 \begin{minipage}{0.49\linewidth}
    \includegraphics[width=1\linewidth]{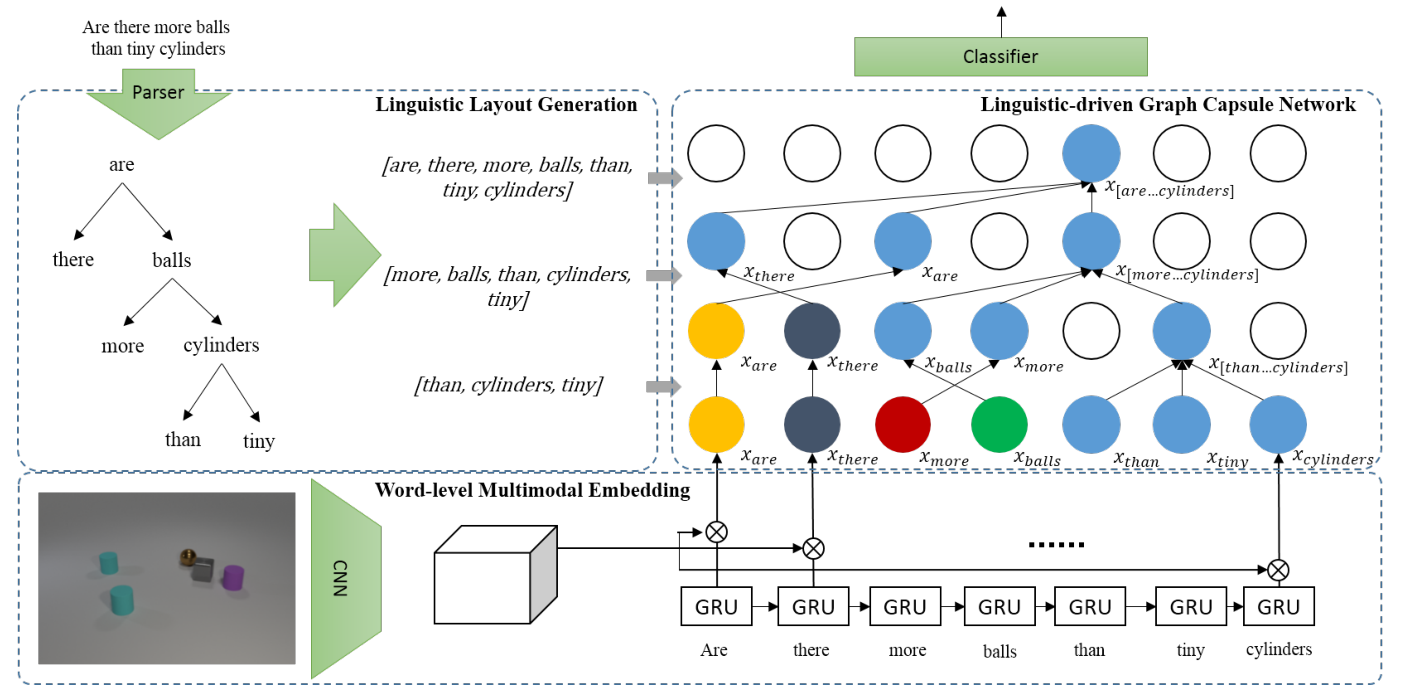}
    \caption{LG-Capsule parses the input question into a tree and fuses image features into the tree nodes. The figure is from reference \cite{cao2020linguistically}.}
    \label{fig:lgcap}
\end{minipage}
\hfill
\begin{minipage}{0.49\linewidth}
    \includegraphics[width=1\linewidth]{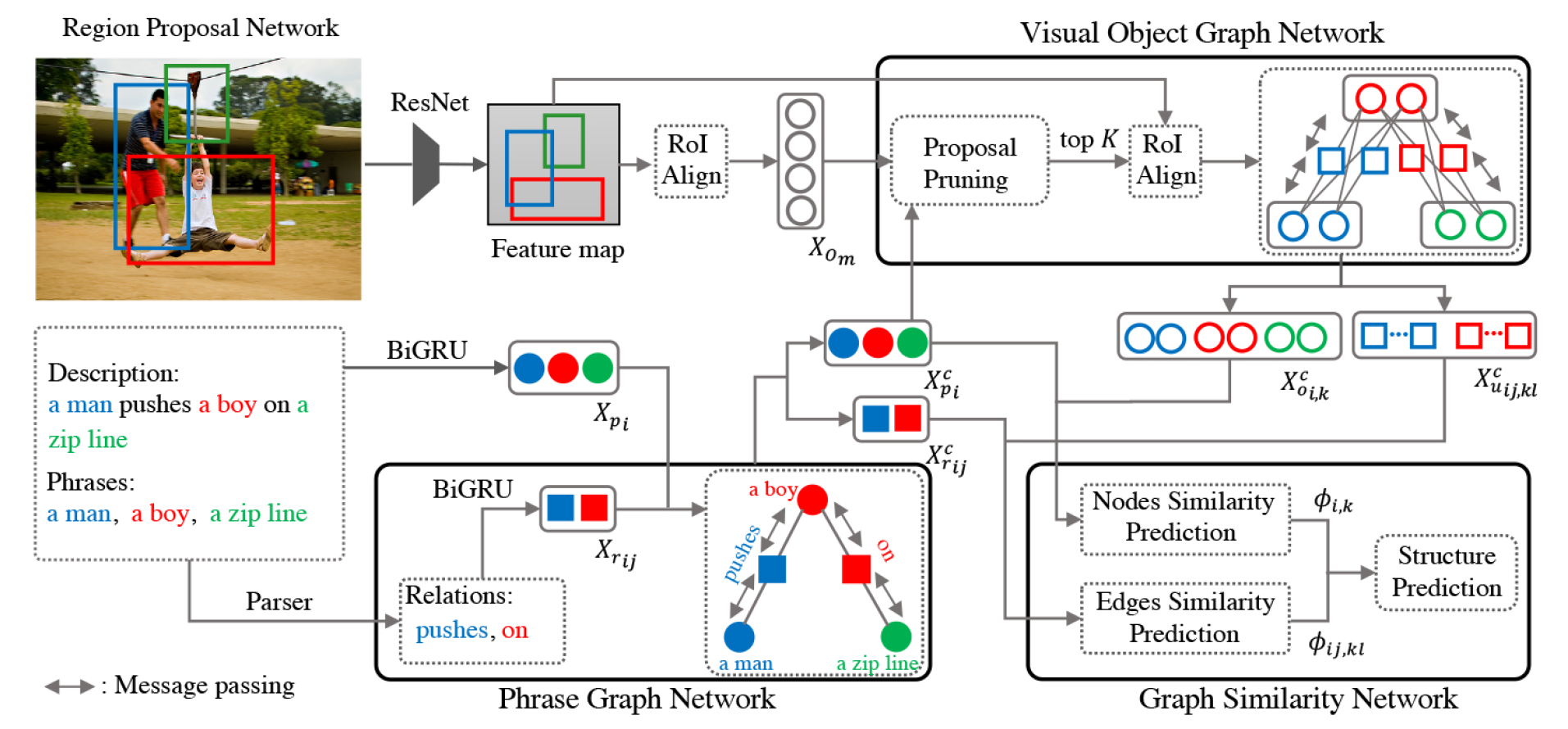}
    \caption{For the visual grounding task, LCMCG constructs a visual object graph and a phrase graph for multimodal inputs and matches visual objects and phrases by graph similarity. The figure is from reference \cite{liu2020learning}.}
    \label{fig:69}
\end{minipage}
\end{figure}

In recent years, research on VQA has explored various end-to-end network architectures. End-to-end trainable LG-Capsule networks \cite{cao2020linguistically} (as shown in Fig. \ref{fig:lgcap}) is proposed, which can incorporate external structured information to retain compositional generalization capabilities while maintaining performance on general tasks. The reasoning process can be performed across multiple graph capsule layers and results in the tree structure indicated by the blue circle.

Indeed, single-modal graph methods are limited in their scope, as they solely construct graphs for the input from a single modality. Consequently, they may fail to explain the reasoning process associated with inputs from other modalities, leaving those aspects unaccounted for in the overall explanation.

\subsection{Multi-modal Graph}
The multi-modal graph contains more diverse types of data and modalities than a single-modal graph, enabling complex reasoning processes through its analysis. 

Liu et al. \cite{liu2020learning} propose a modular graph neural network for vision-based tasks in order to perform reasoning. The model constructs language scene graphs to enhance language representations, and visual scene graphs to refine visual object features. By aligning these two graph representations, the model predicts answers and explains the reasoning process. Additionally, the model generates a sentence based on an image with colored bounding boxes, where it assigns matching colors to the corresponding noun phrases, thus providing a multimodal graph representation.

\begin{figure}[ht]
 \begin{minipage}{0.44\linewidth}
    \includegraphics[width=1\linewidth]{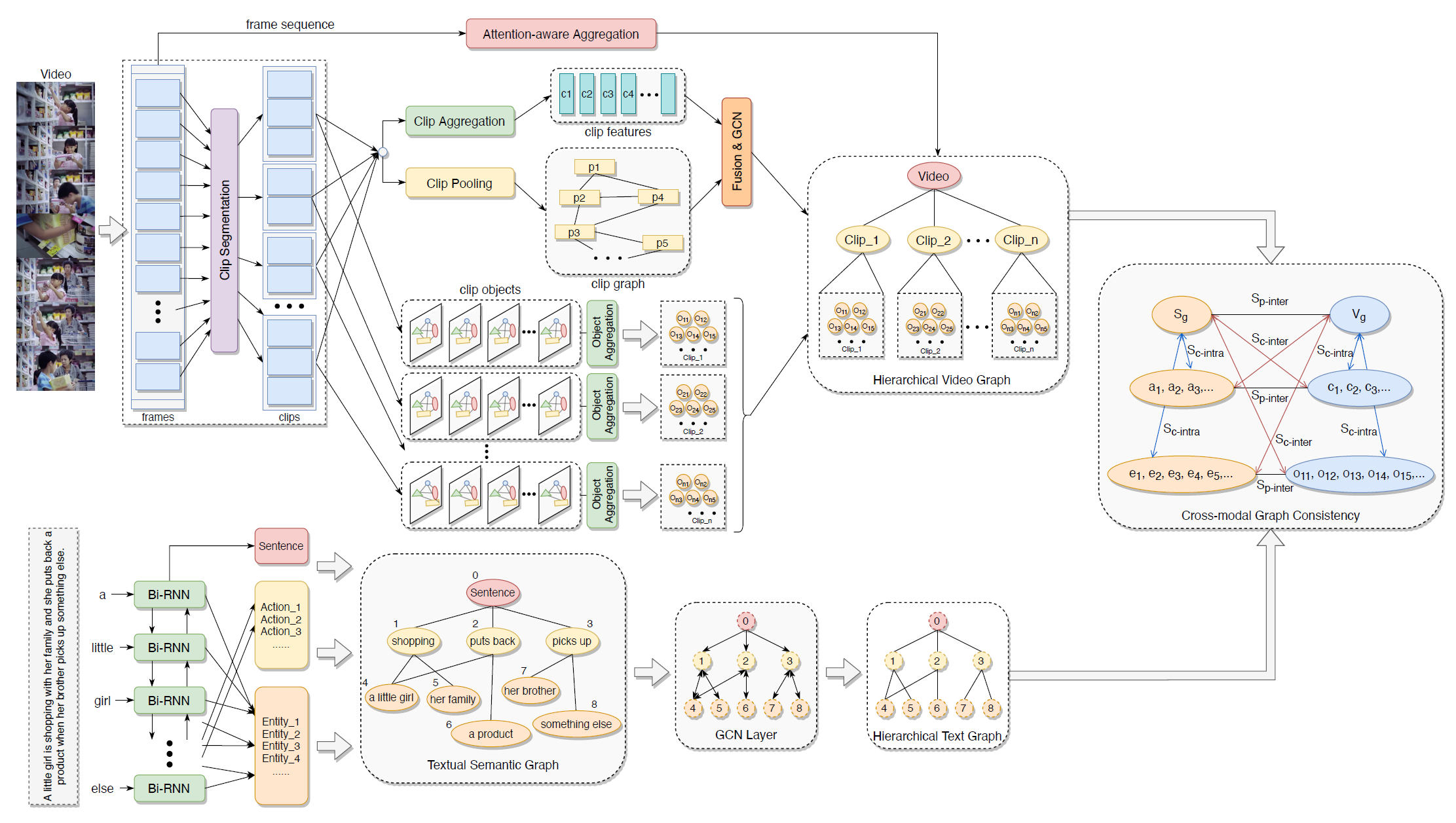}
    \caption{For video-text retrieval task, HCGC constructs a hierarchical video graph and a hierarchical text graph for multimodal queries and computes the cross-modal graph consistency. The figure is from reference \cite{jin2021hierarchical}.}
    \label{fig:71h}
\end{minipage}
\hfill
\begin{minipage}{0.54\linewidth}
    \includegraphics[width=1\linewidth]{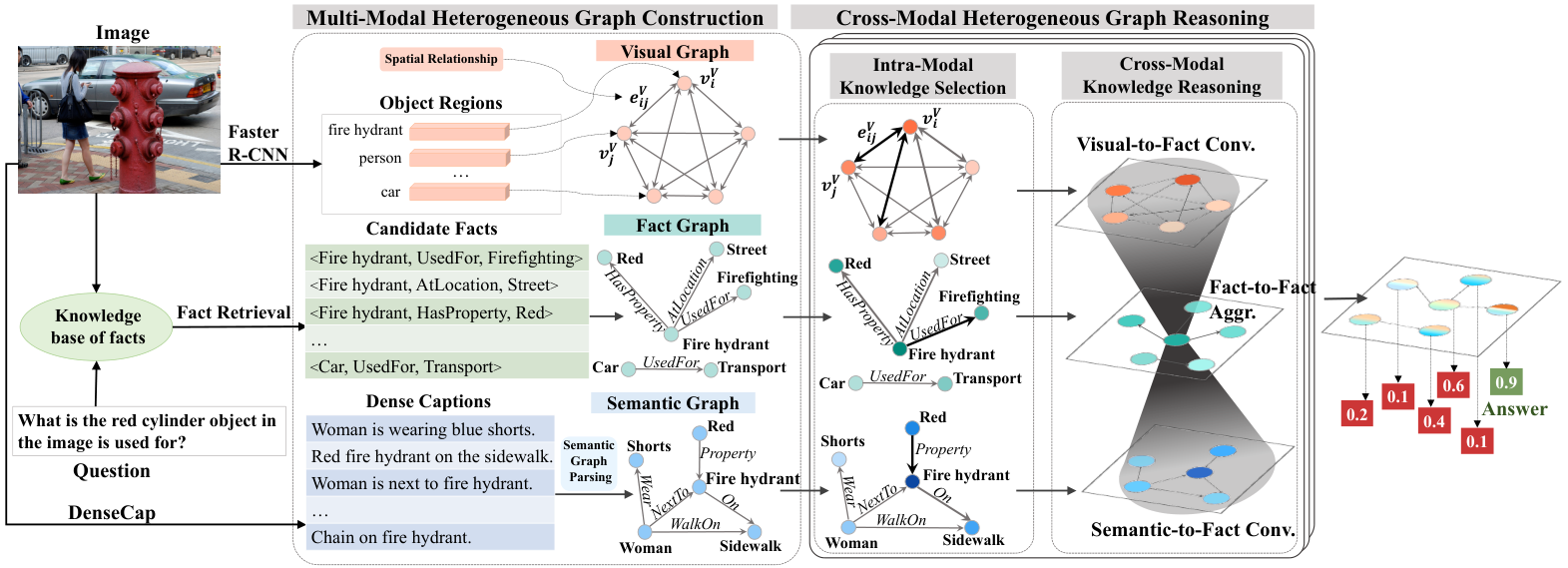}
    \caption{Mucko establishes a visual graph, a fact graph, and a semantic graph based on the input image and question, and then conducts cross-modal heterogenous graph reasoning. The figure is from reference \cite{zhu2021mucko}.}
    \label{fig:70mucko}
\end{minipage}
\end{figure}

To achieve video-text retrieval, Jin et al. \cite{jin2021hierarchical} propose a layered cross-modal graph. As shown in Fig. \ref{fig:71h}, in their model, hierarchical text graphs and hierarchical video graphs are first constructed separately. To learn better matching between the video and text graphs, three types of graph consistencies are designed: inter-graph parallel consistency, inter-graph cross consistency, and intra-graph cross consistency. By exploiting these consistencies, retrieval between text and video is accomplished.

Cheng et al. \cite{ cheng2022cross} utilize cross-modal graphs to achieve image-text retrieval. They employ cross-modal graphs to represent visual and textual information, enabling intra-relation reasoning between regions and words. In their model, both the visual and textual data are encoded and represented as graphs to facilitate matching. During the training process, graph node matching is utilized to facilitate region-word pair alignment, allowing for the learning of fine-grained cross-modal correspondence and inter-relation reasoning.

Jiang et al. \cite{jiang2020kbgn} propose a framework for connecting cross-modal data and constructing a multimodal graph using a Knowledge-Bridge Graph Network (KBGN). To capture the underlying dependence between vision and text modalities, KBGN applies graph structure to bridge multi-modal information. The reasoning begins with text or vision knowledge. They utilize colored bounding boxes to represent objects and arrows, as well as relevant ratios, to explain the process of reasoning, thus providing a multimodal graph for explanation.

As shown in Fig. \ref{fig:70mucko}, Zhu et al. \cite{zhu2021mucko} propose Multi-Layer Cross-Modal Knowledge Reasoning (Mucko) for fact-based VQA, which captures problem-oriented evidence from different modalities. An intra-modal knowledge selection procedure is employed to construct relevant fact graphs, visual graphs, and semantic graphs. Subsequently, in the cross-modal knowledge reasoning procedure, valuable complementary information is selected from the visual graphs and semantic graphs and incorporated into the fact graphs. Finally, based on the fact graphs, inference is performed to form global decisions.

Li et al. \cite{li2022cross} introduce a CMR task namely cross-modal adaptive manipulation (CAM). For CAM, they propose Cross-modal Representation Learning and Relation Reasoning (CRLRR), including two modules: heterogeneous representation learning and cross-modal relation reasoning. Cross-modal relation reasoning identifies and combines the focused attributes and relations in multi-modalities. CRLRR provides explanations by generating visual graphs and language graphs, consisting of nodes, edges, and attributes.

For VQA, Ding et al. \cite{ding2022mukea}  propose a method called Multimodal Knowledge Extraction and Accumulation (MuKEA) for Knowledge-based VQA. It associates visual objects with answers by extracting multimodal information. They implement multi-modal graphs through the use of triplets, where each triplet consists of visual content, a representation of the answer, and the relationship between the two. 
In the visualization of the result of MuKEA, they use the red box in the image to show the head entity and give explanations.

\begin{table}[ht]
\centering
\caption{Categorization of graph explanation methods.}
\begin{tabular}{|c|c|c|}
\hline
\textbf{Category}                   & \textbf{Subcategory}                   & \textbf{Method} \\ \hline
\multirow{8}{*}{Single-modal Graph} & \multirow{4}{*}{Scene Graph}                    & LCGS \cite{norcliffe2018learning}             \\ \cline{3-3} 
                                    &                                                 & ReGAT \cite{li2019relation}           \\ \cline{3-3} 
                                    &                                                 & VSRN \cite{li2019visual2}            \\ \cline{3-3} 
                                    &                                                 & CFR \cite{nguyen2022coarse}            \\ \cline{2-3} 
                                    & \multirow{2}{*}{Knowledge \& Scene Graph} & KM-net \cite{cao2019explainable}             \\ \cline{3-3} 
                                    &                                                 & QAA \cite{zhang2022query}          \\ \cline{2-3} 
                                    & \multirow{2}{*}{Text Graph}                     & UnCoRd \cite{vatashsky2020vqa}     \\ \cline{3-3} 
                                    &                                                 & LG-Capsule \cite{cao2020linguistically}          \\ \hline
\multirow{7}{*}{Multi-modal Graph}  & \multirow{3}{*}{Graph Matching}                 & LCMCG \cite{liu2020learning}            \\ \cline{3-3} 
                                    &                                                 & HCGC \cite{jin2021hierarchical}            \\ \cline{3-3} 
                                    &                                                 & CGMN \cite{cheng2022cross}           \\ \cline{2-3} 
                                    & \multirow{4}{*}{Graph Fusion}                   & KBGN \cite{jiang2020kbgn}           \\ \cline{3-3} 
                                    &                                                 & Mucko \cite{zhu2021mucko}            \\ \cline{3-3} 
                                    &                                                 & CRLRR \cite{li2022cross}          \\ \cline{3-3} 
                                    &                                                 & MuKEA \cite{ding2022mukea}           \\ \hline
\end{tabular}
\label{tab:gra}
\end{table}

\subsection{Discussion}
In CMR tasks, graph explanation methods refer to the methods of constructing graphs that represent entities and their interrelationships within multimodal inputs to facilitate reasoning. 
The graph structure, encompassing nodes and edges, can serve as a tool to explain the underlying reasoning process. 
By leveraging these elements, the reasoning steps and connections between entities can be visually represented and comprehended, offering a clearer understanding of how information is integrated and processed in the context of the given task. 
We summarize the graph explanation methods mentioned above in Table \ref{tab:gra}, where methods of single-modal graphs are categorized by the type of graphs and methods of multi-modal graphs are categorized by the interaction between multiple graphs.
Moreover, graph explanation methods typically share some similar steps as follows:
\begin{itemize}
    \item \textbf{Graph Construction}\quad Graph explanation methods mostly need to extract features from multimodal input and construct graphs that incorporate them. In CMR tasks, developers may extract semantic graphs or phrase structure trees from the text, scene graphs from the image, and clip graphs from the video.

    \item \textbf{Graph Optimization}\quad  Upon the extraction of graphs, graph explanation methods typically further optimize the graph representation. The optimization may include pruning redundant nodes and edges, learning graph features by graph neural networks, and aligning cross-modal graphs. Then, the optimized graph representation can facilitate the reasoning of the final results.

    \item \textbf{Graph Reasoning}\quad After optimizing the graphs, these methods usually conduct graph reasoning by extracting key information from reasoning subgraphs or comparing the structure of cross-modal graphs.
\end{itemize}

The intuitiveness and visualizability of graph explanations provide interpretability and reliability for CMR tasks. However, graph explanation methods may suffer from the following limitations:
\begin{itemize}
    \item \textbf{Information Loss}\quad Converting multimodal inputs into highly structured graphs may cause the loss of semantics and contextual information in the original inputs.
    Especially when using pretrained graph extractors, this problem can be exacerbated due to the domain transfer between the pretrain dataset and the training dataset.


    \item \textbf{Evaluation Criteria}\quad To evaluate the quality of the generated graph explanations, there is currently a lack of universally accepted evaluation criteria for graph alignment. 
    Though many models have achieved visualization of explanatory graphs, the reliability of the generated graphs may be questionable.
\end{itemize}

\section{Symbol Explanation}
Symbol explanation methods symbolize the process of CMR and conduct symbol operations to obtain the reasoning results. The symbols and the process of symbol operations can explain the reasoning process of the CMR model. In general, the research of symbol explanation mainly includes two categories: Logical Inference and Program.

\subsection{Logical Inference}
Logical inference methods involve the extraction of variables from multimodal inputs, along with their attributes and relations. Subsequently, these methods establish their formal systems and conduct logical inferences based on the extracted variables to deduce the reasoning results.

\begin{figure}[ht]
 \begin{minipage}{0.54\linewidth}
    \includegraphics[width=1\linewidth]{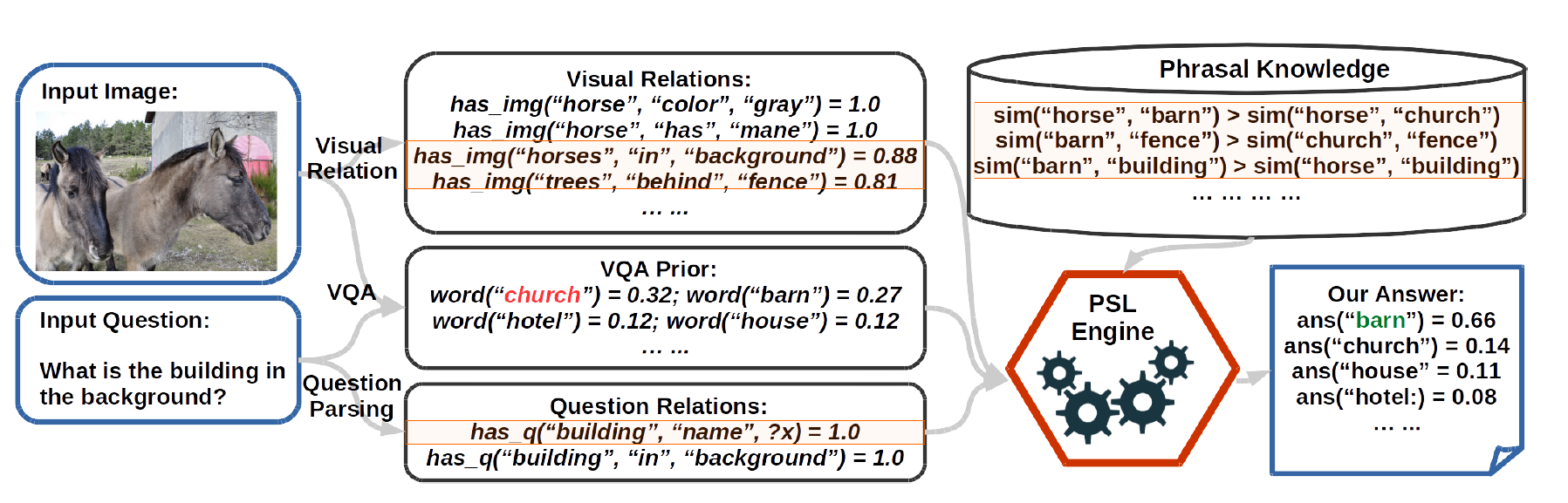}
    \caption{PSLDVQ adopts a PSL-based engine to reason over the inputs: visual relations, question relations, and background ontological knowledge from word2vec and ConceptNet. The figure is from reference \cite{aditya2018explicit}.}
    \label{fig:74A}
\end{minipage}
\hfill
\begin{minipage}{0.44\linewidth}
    \includegraphics[width=1\linewidth]{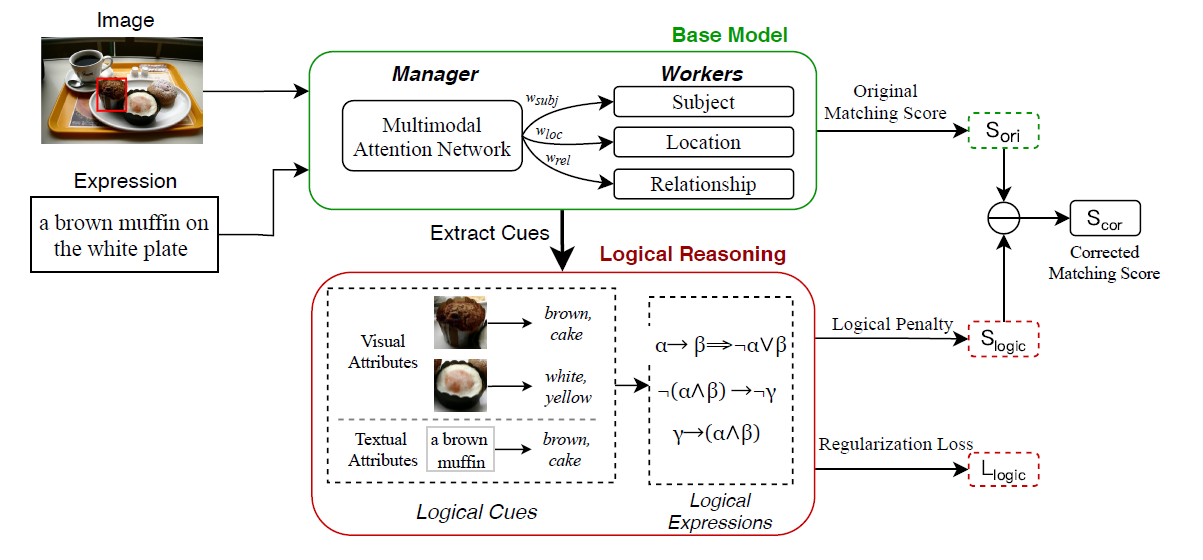}
    \caption{LGREC extracts fine-grained cues as logical symbols and explicitly perform logical reasoning for learning image-expression matching score. The figure is from reference \cite{cheng2021exploring}.}
    \label{fig:80cheng}
\end{minipage}
\end{figure}

Aditya et al. \cite{aditya2018explicit} propose an explicit reasoning layer capable of reasoning and answering questions. As shown in Fig. \ref{fig:74A}, a semantic parser and reasoning component consisting of an augmentation of the Probabilistic Soft Logic (PSL) engine is added to this layer. PSL is a general-purpose probabilistic programming language, which provides syntax that enables users to apply many common modeling techniques \cite{JMLR:v18:15-631}. PSL engine can build a probabilistic graphical model and reason about it to get the prediction.

Suzuki et al \cite{suzuki2019multimodal} propose an unsupervised multimodal logical inference system for symbol explanation. Logic-based representations are used as unified meaning representations for texts and images, which can prove entailment relations between them. The system employs semantic parsing and theorem-proving techniques to effectively process visually complex sentences, thereby enabling visual-textual entailment.

Cheng et al. \cite{cheng2021exploring} introduce two types of logical rules to improve the training and prediction processes, including logical constraints between attribute intersection and matching prediction, as well as implicit logic between visual attribute prediction and subject matching score. These rules were integrated into neural networks using a logic-guided approach to guide the training and prediction, as shown in Fig. \ref{fig:80cheng}.

Ammar et al. \cite{ammar2021spaces} introduce a knowledge graph for logical reasoning. They construct a knowledge graph with multimodal features and populate it with data from multidimensional datasets. The knowledge graph is then fed into a graph-based machine learning algorithm to derive recommendations and insights for better resource allocation and care management. They additionally perform logical inference using rule axioms encoded in a domain ontology.

Liu et al. \cite{liu2023interpretable} propose a logic-based neural model (logicDM) that obtains a series of meaningful logical clauses from the image. In this model, the reasoning process of a task is expressed by logical clauses. Symbolic logic elements are parameterized using neural representations, which facilitate the automatic generation of logic rules consisting of multiple logical clauses.

Several studies have primarily examined the effect of logical connectives on the meaning of sentences in natural language. For visual question answering, Riley and Sridharan \cite{riley2019integrating} utilize incomplete commonsense domain knowledge and decision tree induction components. In the context of decision tree classifiers, every node is associated with a specific feature value, and its child nodes correspond to diverse answers. By leveraging active nodes in the decision tree classification path, a deeper and more sophisticated explanation of the classification outcomes can be provided.

The consideration of logical connectives and their impact on the semantic interpretation of natural language sentences is imperative. Lens Of Logic (LOL) model is an end-to-end model equipped with specialized attention modules, which can answer questions by comprehending the logical connectives within them \cite{gokhale2020vqa}. Their model primarily hinges upon the utilization of two extensive datasets, namely VQA-Compose and VQA-Supplement, which are repositories of copious amounts of logically and systematically composed binary questions.

The employment of logical inference methods presents a promising avenue for facilitating a more intuitive and interpretable reasoning process. Nevertheless, the construction and refinement of such methods have proven to be a formidable challenge, primarily attributed to the heightened intricacy and abstractness of CMR tasks in contrast to single-modal tasks. Furthermore, the absence of suitable datasets for extracting multifaceted logical connections exacerbates the already intricate task.

\subsection{Program}
Program methods are commonly employed for symbolism analysis, enabling the generation of multiple modular implementations for program execution. Each of these modules is designed to accept inputs, autonomously produce their own outputs, and operate in a collaborative manner. The computation results of modules, along with the integrated computation process of the entire program, can serve as explanations for the reasoning process, hence augmenting the interpretability of CMR methods.

\begin{figure}[ht]
 \begin{minipage}{0.49\linewidth}
    \includegraphics[width=1\linewidth]{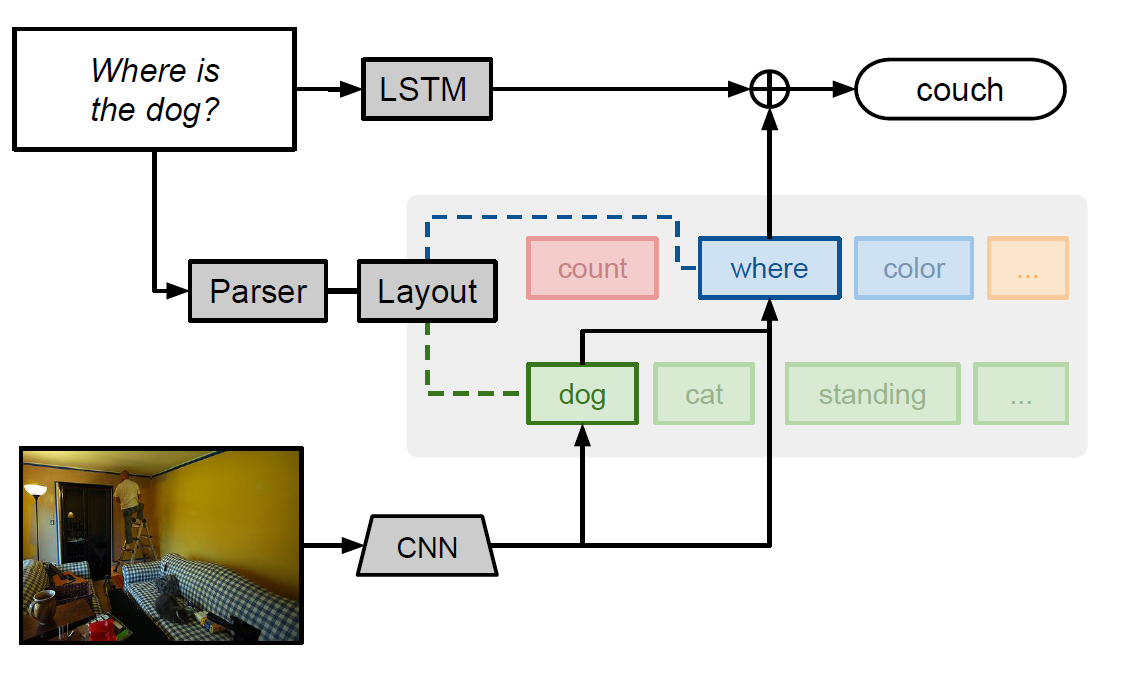}
    \caption{Neural Module Networks (NMN) uses a natural language parser to dynamically lay out a deep network composed of reusable modules. The figure is from reference \cite{andreas2016neural}.}
    \label{fig: and16}
\end{minipage}
\hfill
\begin{minipage}{0.49\linewidth}
    \includegraphics[width=1\linewidth]{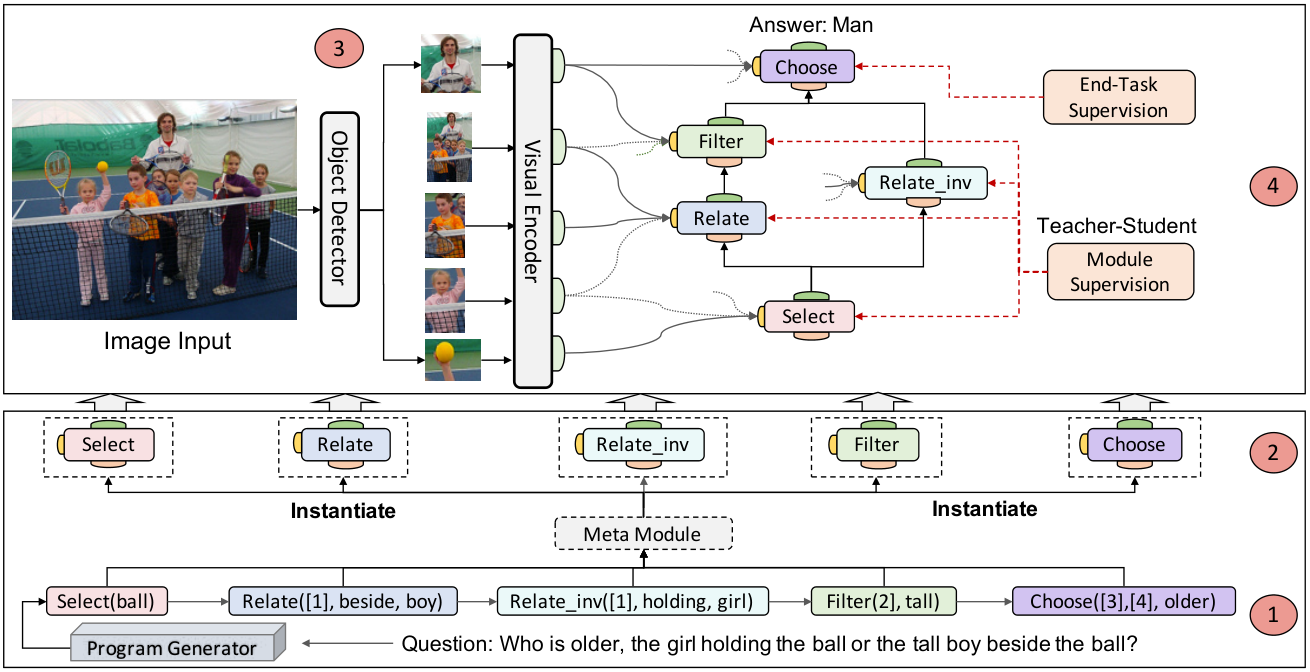}
    \caption{Meta Module Network (MMN) instantiate modules from the meta module based on the input function recipes (specifications), every instance module has shared parameterization. The figure is from reference \cite{chen2021meta}.}
    \label{fig:mmn}
\end{minipage}
\end{figure}

Drawing upon program methods, novel neural network architectures are proposed. Previous modular networks first analyze the problem and then predict a sequence of pre-defined program modules that are linked together to make a prediction. As shown in Fig. \ref{fig: and16}, Andreas et al. \cite{andreas2016neural} have posited neural module networks (NMN) as a comprehensive architecture for the discrete components of diverse, jointly-trained neural modules within complex network structures. The approach employed by the researchers involves the deconstruction of natural language queries into a series of linguistic substructures, which are subsequently implemented within the neural network model.

Previous NMN implementations rely on brittle off-the-shelf parsers and are restricted to the module configurations proposed by these parsers rather than learning them from data. To tackle these problems, Hu et al. \cite{hu2017learning} further develop End-to-End Module Networks (N2NMN), which learn to reason by directly predicting instance-specific network layouts without the aid of a parser. This model can learn to generate network structures while simultaneously learning network parameters. Consequently, N2NMNs inherit the interpretability of NMN while significantly improving the accuracy of reasoning.


Yi et al. \cite{yi2018neural} propose models with some structural similarities. Their neural-symbolic visual question answering (NS-VQA) mechanism initially extracts a structural scene representation from the image and subsequently retrieves a program trace from the inquiry. The program is subsequently implemented on the scene representation in order to derive an answer.

Hu et al.\cite{hu2018explainable} contend that the module network model overlooks the significance of module arrangement. To achieve superior accuracy, they suggest training a placement policy with supervised module placement. Their proposed method, the Stack Neural Module Network (SNMN) obviates the need for layout supervision and replaces the layout graph with a stack-based data structure. Instead of employing discrete selections for module layout, this study adopts an approach whereby the layout is rendered as soft and continuous, thus enabling full differentiability of the model-based optimizations through the use of gradient descent. 

Recently, Hsu et al. \cite{hsu2023ns3d} introduce neural module networks to the grounding of 3D objects and relations. They propose NS3D, which translates language into programs with hierarchical structures by leveraging large language-to-code models. Notably, NS3D extends prior neuro-symbolic visual reasoning methods by introducing functional modules that effectively reason about high-arity relations (i.e., relations among more than two objects), key in disambiguating objects in complex 3D scenes

In NMN-based methods, each module has its independent parameterization, which may hinder the scalability and generalizability of modules.
To tackle this problem, Chen et al. \cite{chen2021meta} propose Meta Module Network (MMN) centered on a novel meta module, which can take in function recipes and morph into diverse instance modules dynamically, as shown in Fig. \ref{fig:mmn}. 
The instance modules are then woven into an execution graph for complex reasoning, inheriting the strong explainability and compositionality of NMN. 
The parameters of instance modules are inherited from the central meta module, which promises better scalability. Meanwhile, unseen functions can be readily represented in the embedding space based on their structural similarity with previously observed ones, which ensures better generalizability. 

Zhao et al.\cite{zhao2021proto} follow MMN and accord greater importance to program executors by formulating program-guided tasks that necessitate the agent to execute given programs while conditioned on task specifications. They propose the Program-guided Transformer (ProTo), which combines the robust representation capability of transformers with symbolic program control flow. To further enhance its effectiveness, ProTo adopts efficient attention mechanisms to separately leverage program semantics and explicit structures. 

\begin{figure}[ht]
 \begin{minipage}{0.49\linewidth}
    \includegraphics[width=0.95\linewidth]{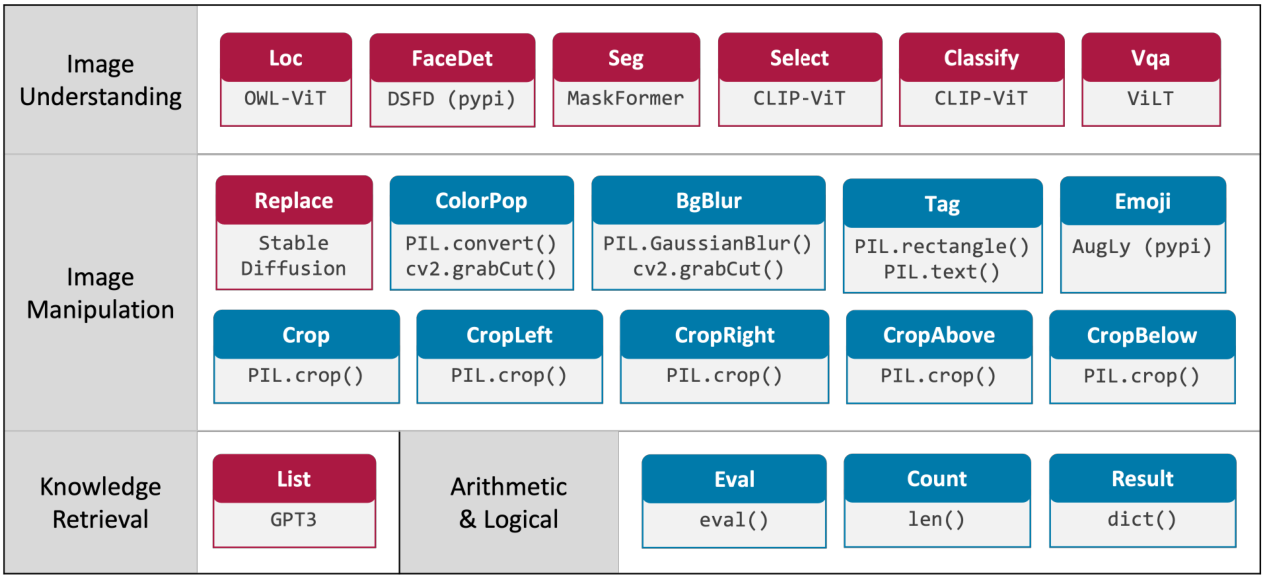}
    \caption{VisProg adopts both trained neural modules (colored red) and Python subroutine modules (colored blue). The figure is from reference \cite{gupta2023visual}.}
    \label{fig:visprog}
\end{minipage}
\hfill
\begin{minipage}{0.49\linewidth}
    \includegraphics[width=1\linewidth]{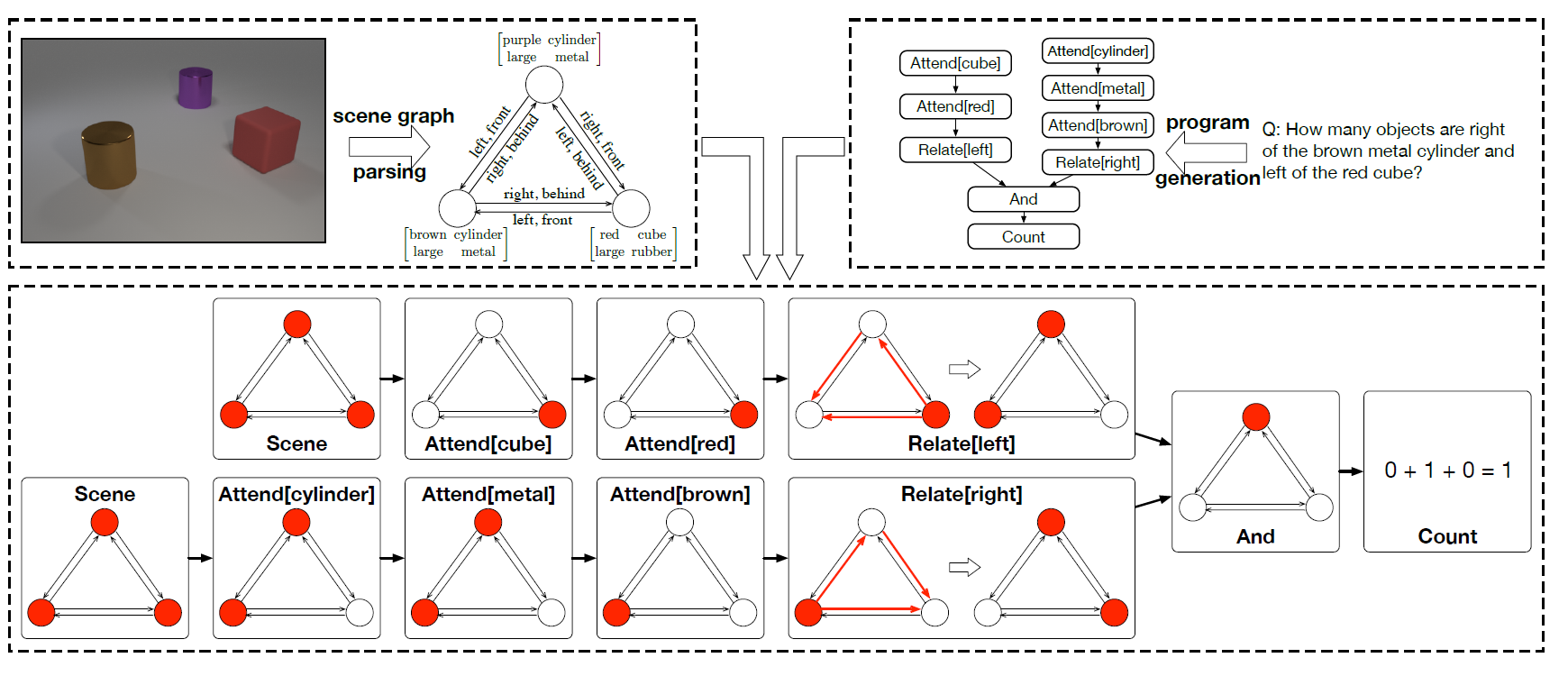}
    \caption{XNMs parse the image into a scene graph, parse the question into a module program, and then run the program on the scene graph. The figure is from reference \cite{shi2019explainable}.}
    \label{fig: XNM}
\end{minipage}
\end{figure}

Different from the above methods, Gupta and Kembhavi \cite{gupta2023visual} recently propose a visual program to handle compositional visual reasoning. They harness the contextual learning capabilities of large language models to generate visual programs tailored for visual tasks, utilizing natural language instructions. Specifically, GPT-3 \cite{brown2020language} is prompted to inform it about the input and output types and functionalities of each module. VisProg includes 20 hybrid modules, consisting of 8 neural modules and 12 Python subroutine modules, as shown in Fig. \ref{fig:visprog}. The program execution is handled by the interpreter, which sequentially traverses the program invoking the correct module with the specified inputs. The inputs and outputs of each module are visually summarised. The interpreter then compiles all summaries of program steps into a visualized symbol explanation.
%

\begin{table}[ht]
\centering
\caption{Categorization of symbol explanation methods.}
\begin{tabular}{|c|c|c|}
\hline
\textbf{Category}                           & \textbf{Subcategory}                              & \textbf{Method}                                            \\ \hline
\multirow{7}{*}{Logical Inference} & Probabilistic Soft Logic                 & \multicolumn{1}{c|}{PSLDVQ \cite{aditya2018explicit}} \\ \cline{2-3} 
                                   & \multirow{4}{*}{First-Order Logic}       & MLIS \cite{suzuki2019multimodal}                      \\ \cline{3-3} 
                                   &                                          & LGREC \cite{cheng2021exploring}  \\ \cline{3-3} 
                                   &                                          & SPACES \cite{ammar2021spaces}                         \\ \cline{3-3} 
                                   &                                          & LogicDM \cite{liu2023interpretable}                   \\ \cline{2-3} 
                                   & Non-monotonic Logic & NLRIL \cite{riley2019integrating}                     \\ \cline{2-3} 
                                   & Propositional Logic & LOL \cite{gokhale2020vqa}                             \\ \hline
\multirow{8}{*}{Program}           & \multirow{5}{*}{Specialized Neural Modules} & NMN \cite{andreas2016neural}        \\ \cline{3-3} 
                                   &                                             & N2NMN \cite{hu2017learning}         \\ \cline{3-3} 
                                   &                                             & NS-VQA \cite{yi2018neural}          \\ \cline{3-3} 
                                   &                                             & SNMN \cite{hu2018explainable}       \\ \cline{3-3} 
                                   &                                             & NS3D \cite{hsu2023ns3d}       \\ \cline{2-3} 
                                   & \multirow{2}{*}{Meta Neural Module}         & MMN \cite{chen2021meta}             \\ \cline{3-3} 
                                   &                                             & ProTo \cite{zhao2021proto}          \\ \cline{2-3} 
                                   & Hybrid Modules                              & VisProg \cite{gupta2023visual}      \\ \hline
\end{tabular}
\label{tab:sym}
\end{table}

\subsection{Discussion}
Methods of symbol explanation conduct symbol deduction to facilitate cross-modal reasoning, whereby the deducing processes themselves serve as symbol explanations.
Therefore, the provided explanations can consistently reflect the cross-modal reasoning processes of models.
We summarize the symbol explanation methods mentioned above in Table \ref{tab:sym}. Additionally, we further classify methods of logical inference based on the types of formal systems \cite{smullyan1961theory}, and classify methods of program based on the types of the program modules.

Moreover, it can be observed that the abovementioned approaches of symbol explanation exhibit a notable similarity in their reasoning procedures. 
In the field of symbolic explanation, some commonly employed steps in cross-modal reasoning can be summarized as follows:
\begin{itemize}
    \item \textbf{Problem Symbolization}\quad Methods of which the specific reasoning problem depends on input instructions \cite{aditya2018explicit, andreas2016neural, hu2017learning, gupta2023visual, hsu2023ns3d, zhao2021proto} usually begin by symbolizing the textual instruction to obtain a formalized statement or program.

    \item \textbf{Extraction of Variable Attribution and Relation}\quad To facilitate symbol deduction, methods of symbol explanation typically extract the attributions and relations of variables to compute the results of axiom expressions or functions.

    \item \textbf{Symbol Deduction}\quad Symbol methods finally integrate the results of all axiom expressions or functions to deduce the final result of the complete statement or program. The values of axioms and the deduction processes can serve as symbol explanations.
\end{itemize}
%

%

Despite the consistency between the explanations and the reasoning processes, symbol methods typically suffer from some common limitations:
\begin{itemize}
\item \textbf{Limited Robustness}\quad Symbolizing textual instructions is a crucial step for many symbol methods.
However, these methods typically conduct experiments on datasets where textual instructions are constructed by predefined processes without noise. 
Therefore, the performance of problem symbolization may be unstable when user instructions in real scenarios are inputted, including those out-of-distribution expressions.
A potential solution for enhancing the robustness is to leverage the powerful generalization capability of large-scale pre-trained language models \cite{brown2020language, du2022glm, touvron2023llama}, as Gupta and Kembhavi \cite{gupta2023visual} recently have explored.
\item \textbf{Error Accumulation}\quad Symbol methods typically integrate the results of all axiom expressions or functions to deduce the final result. Since the error of parent nodes/axioms can propagate to child nodes/axioms, the reasoning error of some axioms can be accumulated and significantly affect the final result.
To alleviate this problem, some methods \cite{hu2018explainable, zhao2021proto, liu2023interpretable} conduct the deduction by utilizing the feature vectors of axioms instead of the scalar values (e.g., ``yes" and ``no").
However, the error accumulation problem is still not completely addressed, especially considering the common performance gap compared to other reasoning methods.
\item \textbf{Limited User-friendliness}\quad Though the symbol deduction process can be a consistent explanation of the cross-modal reasoning process, it requires corresponding knowledge to understand the meaning of symbols.
Therefore, symbol explanation can be difficult to understand by a wide range of general users without corresponding knowledge.
A potential solution is to combine more user-friendly explanations based on the symbol explanation.
\end{itemize}

\section{Multimodal Explanation}
\label{sec:mul}

In contrast to conventional single-modal explanation methods, which solely explain the CMR process by single modality, multimodal explanation methods incorporate multiple modalities simultaneously to facilitate a more comprehensive and holistic interpretation of the CMR process. Multimodal explanation methods can be classified into two groups, namely Independent Explanations and Joint Explanation, based on whether the multiple modalities are utilized to form separate explanations or integrated to form a unified explanation.

\subsection{Independent Explanations}

\begin{figure}[ht]
 \begin{minipage}{0.44\linewidth}
    \includegraphics[width=1\linewidth]{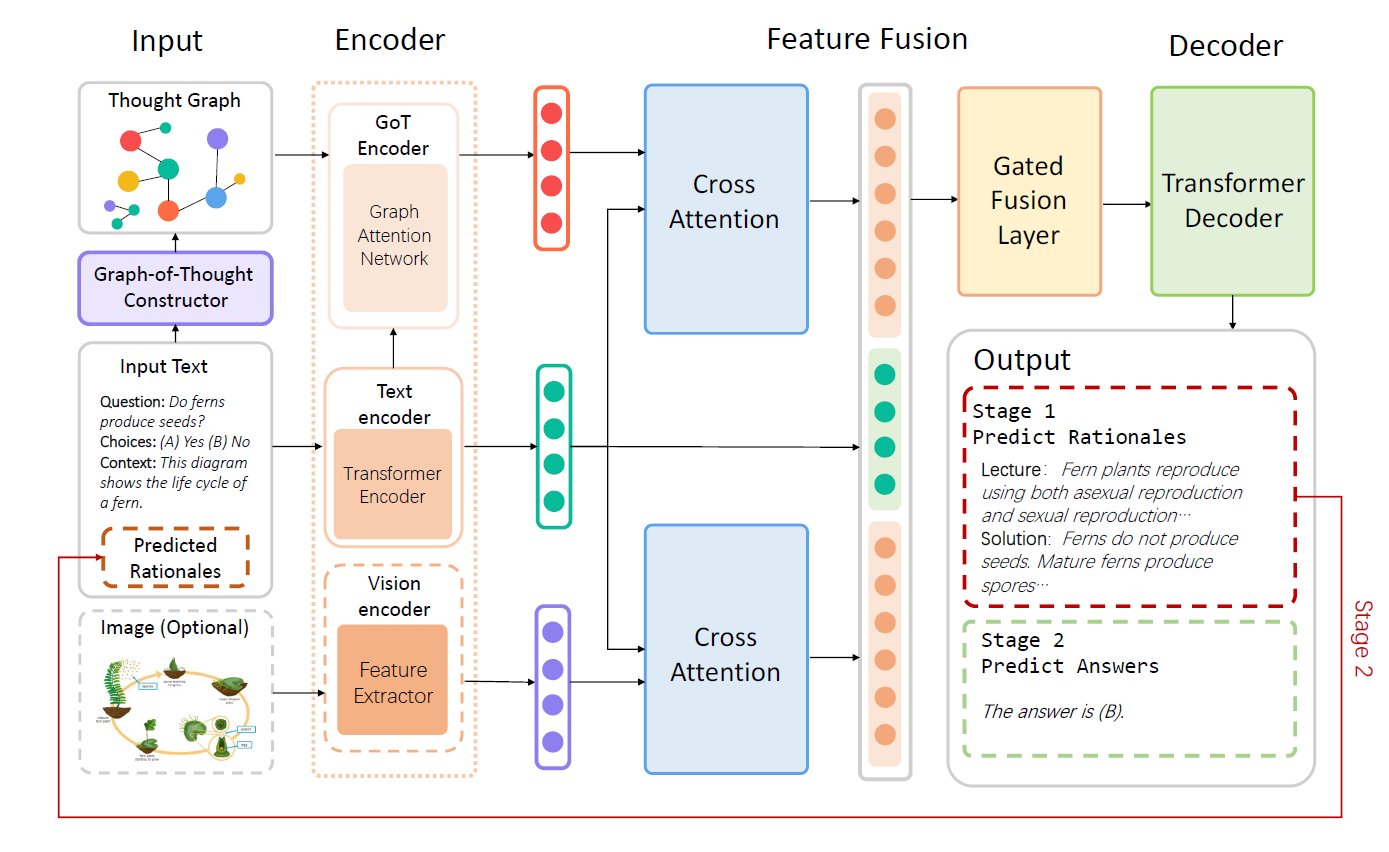}
    \caption{Graph-of-Thought constructs the thought graph corresponding to the input text and then generates the textual rationales based on multimodal input and the thought graph.  The figure is from reference \cite{yao2023beyond}.}
    \label{fig: yao2023}
\end{minipage}
\hfill
\begin{minipage}{0.54\linewidth}
    \includegraphics[width=1\linewidth]{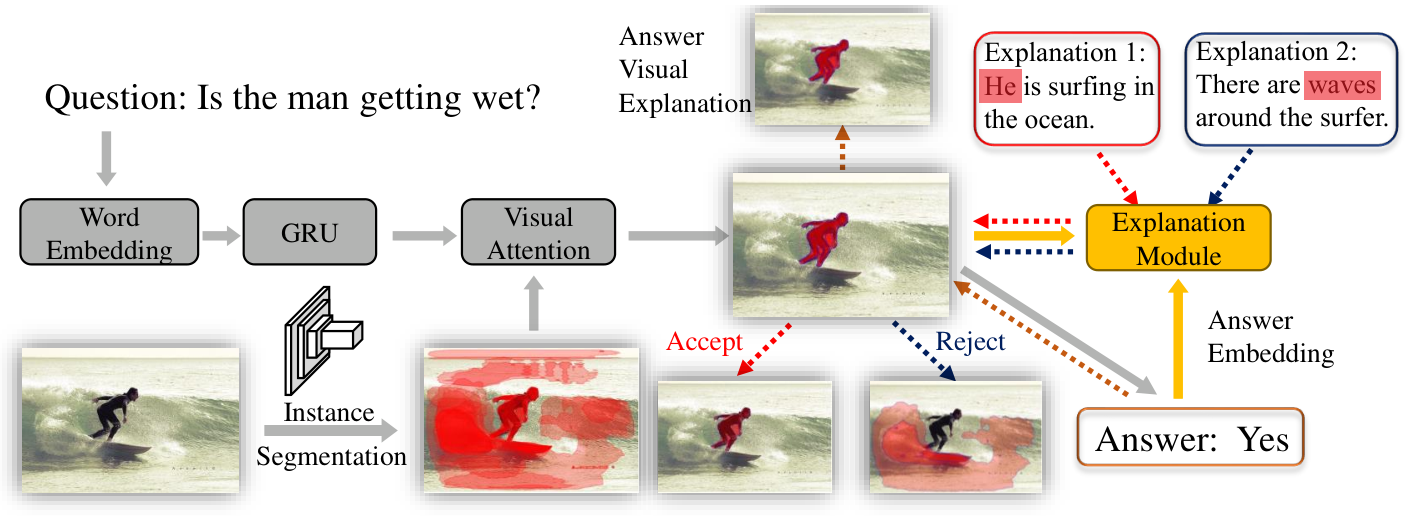}
    \caption{Faithful Multimodal Explanation (FME) first generates the textual explanation by LSTM with attention to visual regions and then ground explanation words with relevant visual regions by attention weights. The figure is from reference \cite{wu2019faithful}.}
    \label{fig:93overview}
\end{minipage}
\end{figure}

Yao et al. \cite{yao2023beyond} propose a multi-modal reasoning framework inspired by CoT and incorporating human thought processes, known as Graph-of-Thought (GoT) reasoning. This approach models the human thinking process as a graph.  The model independently generates rationales for each modality in the rationale generation stage based on the input multi-modal information (refer to Fig. \ref{fig: yao2023}). In the answer generation stage, these rationales, combined with the input multi-modal information, are used as input to generate predictions and multi-modal explanations. 

In the context of separately handling information from different modalities, there are multiple ways to integrate the outputs of different modalities for answer prediction. Shi et al. \cite{shi2019explainable} propose eXplainable and eXplicit Neural Modules (XNMs) based on scene graphs. This model processes the image to obtain scene graphs and then processes the textual information to obtain programs. Finally, the program is executed on the scene graphs to predict the results and provide explanations, as shown in Fig. \ref{fig: XNM}.

Park et al. \cite{park2018multimodal} propose a multimodal explanation system. To justify the reasoning behind decisions using natural language and provide corresponding evidence, they introduce the Pointing and Justification Model (PJ-X). This model consists of two components: an answering model predicts the answer, while a multimodal explanation model generates an explanation based on the answer and multimodal information. PJ-X is enabled to predict the answer and generate rationales that both provide textual evidence and point to visual evidence.  


\subsection{Joint Explanation}
In contrast to simultaneously generating independent explanations of multiple modalities, Joint explanation methods aim to combine multiple modalities to form a unified explanation of improved interpretability.

Zellers et al. \cite{zellers2019recognition} aims to improve previous object recognition approaches by introducing a task called Visual Commonsense Reasoning (VCR) and proposing a corresponding dataset. Building upon this, they propose a reasoning engine, Recognition to Cognition Networks (R2C), to model the necessary layered inference for grounding, contextualization, and reasoning. Based on R2C, they insert images with labels and colors into textual explanations.

As shown in Fig. \ref{fig:93overview}, Wu and Mooney \cite{wu2019faithful} propose Faithful Multimodal Explanation (FME). Firstly, they segment the image into multiple regions and extract the most salient ones. They then utilize a pre-trained VQA module for answer prediction. Finally, the model learns how to embed the question, answer, and VQA attention features to generate textual explanations. Their multimodal explanation highlights relevant image regions together with a textual explanation with corresponding words in the same color.

\begin{figure}[ht]
 \begin{minipage}{0.49\linewidth}
    \includegraphics[width=1\linewidth]{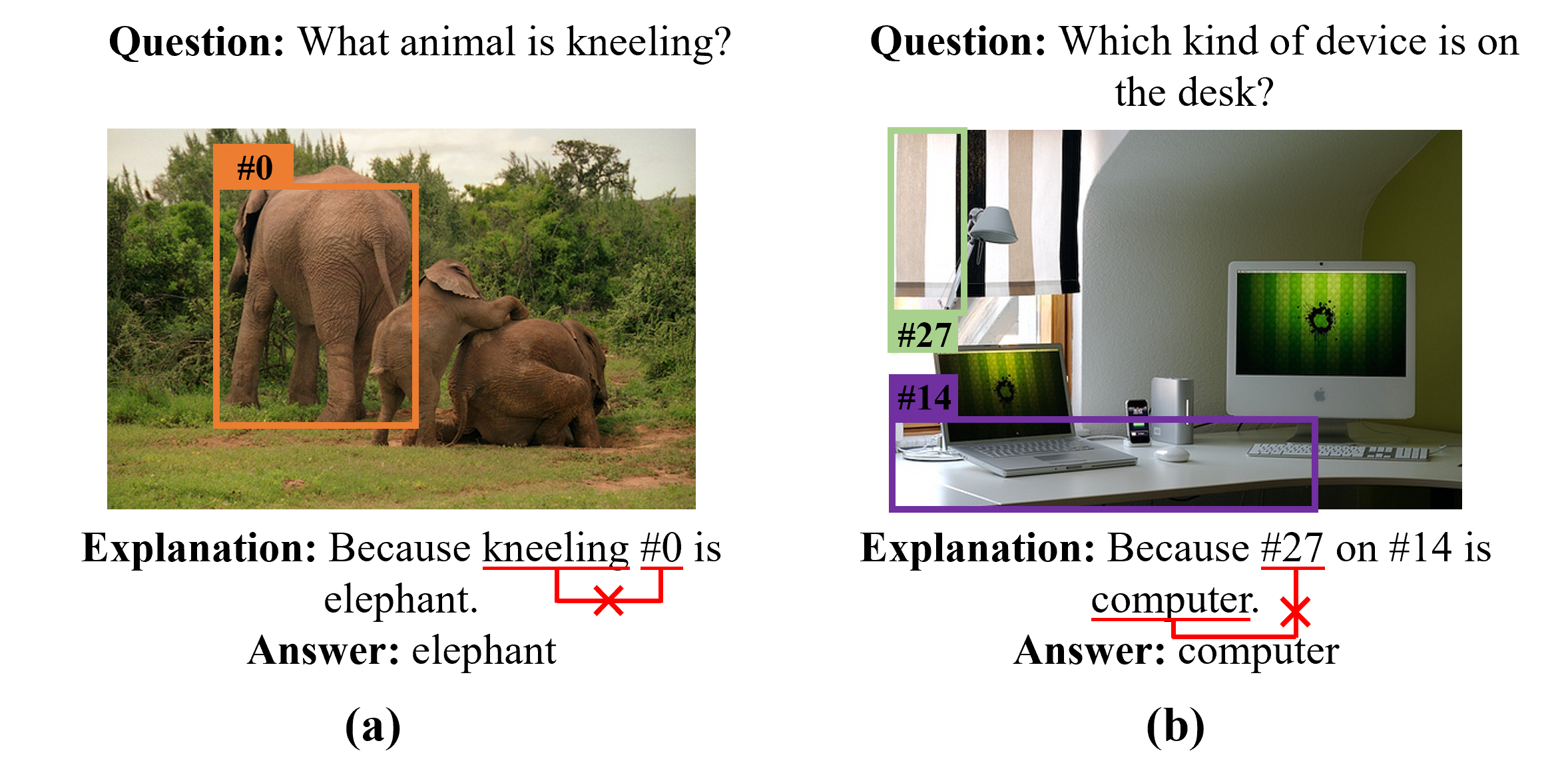}
    \caption{Semantic inconsistency between different modalities in explanations generated by REX \cite{chen2022rex}: (a) \#0 is not kneeling; (b) \#27 is not computer.}
    \label{fig:incons}
\end{minipage}
\hfill
\begin{minipage}{0.49\linewidth}
    \includegraphics[width=1\linewidth]{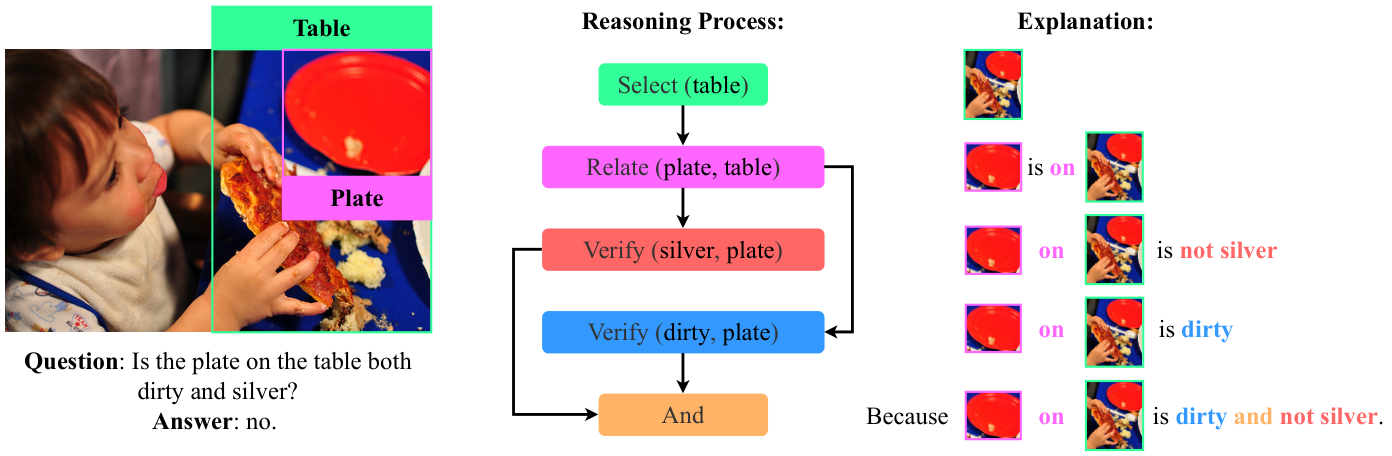}
    \caption{REX proposes a multimodal explanation of reasoning processes for visual question answering, which utilizes textual description and grounds key objects in the image. The figure is from reference \cite{chen2022rex}.}
    \label{fig:17-1}
\end{minipage}
\end{figure}

To achieve a joint explanation, Chen and Zhao \cite{chen2022rex}  research visual reasoning from both the data and model perspectives. On the data side, they proposed a reasoning-aware and explanatory dataset called VQA-REX, along with a Reasoning-aware and Grounded Explanation (REX) benchmark. On the model side, they propose a corresponding explanation generation method that combines the relevant components from both modalities and generates explanations based on their relationships. From Fig. \ref{fig:17-1}, the explanation consists of text and images and provides strong interpretability.

\begin{table}[ht]
\centering
\caption{Categorization of multimodal explanation methods.}
\begin{tabular}{|c|c|c|}
\hline
\textbf{Category}                         & \textbf{Modality Combination} & \textbf{Method} \\ \hline
\multirow{3}{*}{Independent Explanations} & Graph+Text                    & GoT \cite{yao2023beyond}             \\ \cline{2-3} 
                                          & Graph+Symbol                  & XNMs \cite{shi2019explainable}            \\ \cline{2-3} 
                                          & \multirow{4}{*}{Visual+Text}  & PJ-X \cite{park2018multimodal}            \\ \cline{1-1} \cline{3-3} 
\multirow{3}{*}{Joint Explanation}        &                               & R2C \cite{zellers2019recognition}             \\ \cline{3-3} 
                                          &                               & FME \cite{wu2019faithful}             \\ \cline{3-3} 
                                          &                               & REX \cite{chen2022rex}             \\ \hline
\end{tabular}
\label{tab:mul}
\end{table}

\subsection{Discussion}

Due to the complexity of involving different modalities, research concerning multimodal explanations remains comparatively limited.
The objective of multimodal explanation is to generate easily understandable explanations for the CMR processes by combining different modalities, such as texts and images. 
These explanations are typically presented through natural language descriptions, visualizations, and other means. 
We summarize the multimodal explanation methods mentioned above in Table \ref{tab:mul}.
Furthermore, some common similarities shared by multimodal explanation methods are summarized as follows:
\begin{itemize}
    \item \textbf{Multimodal Generative Modules}\quad To generate multimodal explanations, these methods usually involve generative modules of multiple modalities, such as language generator, graph constructor, and visual attention.

    \item \textbf{Interaction between Modalities}\quad While using multiple modalities in explanations, the majority of multimodal explanation methods take into account the interaction between modalities. Some methods generate the explanation of one modality based on the generated explanation of another modality \cite{yao2023beyond, shi2019explainable, park2018multimodal}. Some methods construct relevance between textual words and visual regions in explanations \cite{chen2022rex, wu2019faithful, zellers2019recognition}.

    \item \textbf{Multiple Perspectives on Reasoning Process}\quad By harnessing the advantages derived from multiple modalities in the generated explanations, multimodal explanation methods offer diverse perspectives on the CMR process. 
    Consequently, these approaches provide more comprehensive insights into the CMR process, significantly enhancing the interpretability of CMR models.
\end{itemize}

Despite the multi-perspective interpretability of multimodal explanation methods, these methods and the research can suffer from some common limitations:
\begin{itemize}
    \item \textbf{Modality Inconsistency}\quad While combining multiple modalities in explanations, the correlation between modalities can be weak in some methods, leading to inconsistent explanations. For example, the grounded visual objects may be inconsistent with the textual description in the generated explanations, as shown in Fig. \ref{fig:incons}. 
    Such semantic inconsistency can impede the credibility and reliability of the I-CMR models.

    \item \textbf{Complexity}\quad The utilization of multiple modalities in the explanations can also improve the complexity of accurately modeling the reasoning process.
    Especially for CMR tasks where the supervision of the explanations is missing, more modalities involved may lead to more inaccurate explanations.

    \item \textbf{Task Diversity}\quad Existing research on the multimodal explanation for CMR predominantly focuses on the VQA task, including the previously mentioned works. 
    There exists a promising avenue for further exploration of multimodal explanation by considering a broader spectrum of tasks, including visual grounding, vision-and-language navigation, and cross-modal retrieval. 
\end{itemize}

\section{Datasets}
In this section, we introduce prevalent CMR datasets annotated with explanations or data that can be utilized for explanatory purposes.
While some I-CMR models can be trained on generic CMR datasets without the necessity for annotated explanations, the quality of the provided explanations may be doubtful due to a lack of appropriate evaluation.
With the annotations for explanations, interpretable models can be supervised to improve the quality of the provided explanations. The quality of these explanations can then be critically evaluated and objectively compared with those derived from other models.

\textbf{Visual Genome} \cite{krishna2017visual} is a vision-and-language dataset with over 100K images, which can be adopted for CMR tasks such as phrase grounding and visual question answering (VQA). Visual Genome annotates visual objects, object attributes, scene graphs, region descriptions, and visual question answers for images. 
Therefore, the annotated visual objects can be used to construct importance maps in visual explanation methods for I-CMR.
The scene graphs can be utilized by graph explanation methods.
The region descriptions can be adopted by textual explanation methods to describe the semantics of regions.

\textbf{GQA} \cite{hudson2019gqa} is a VQA dataset based on Visual Genome, which contains 22M diverse reasoning questions. 
Every sample in GQA contains an image, an image-related question, and the answer.
Moreover, GQA annotates functional programs to represent the semantics of questions, which can be utilized by symbol explanation methods for I-CMR.
Additionally, GQA inherits the annotations of scene graphs and visual objects for images in Visual Genome, which can be utilized by graph explanation methods and visual explanation methods.

\textbf{GQA-REX} \cite{chen2022rex} is a dataset with more than 1M multi-modal explanations, constructed based on the balanced training and validation sets of GQA. \cite{hudson2019gqa}. 
GQA-REX dataset comprises a training set, which serves as the basis for optimizing VQA models, and a validation set, used to assess the performance of explanation generation.
Hence, the utilization of GQA-REX enables the evaluation of multimodal explanations from various perspectives, encompassing reasoning performance, explanation quality, visual grounding, and attribute recognition.

\textbf{FVQA} \cite{wang2017fvqa} is a VQA dataset that provides supporting facts for each question-answer pair as supplementary information, which sets it apart from previous VQA datasets.
The addition of supporting facts in FVQA aims to enhance the evaluation of VQA models. A sample in FVQA consists of an image, a question, an answer, and a supporting fact. In total, FVQA contains 2190 images and 5826 questions (corresponding to 4216 unique facts).
The annotated supporting facts can be utilized to supervise the generation of textual explanation methods.

\textbf{VQA-E} \cite{li2018vqa} is a dataset with 108,325 images, 269,786 question-answer pairs, and relevant explanations, created by exploiting the available captions and generating an explanation for each image-question-answer triple from VQA v2 \cite{goyal2017making}.
VQA-E provides insightful information that can explain answers compared with the traditional VQA task.
By annotating related image captions about questions, VQA-E points out where to look for the answer, which can be utilized by textual explanation methods. 

\textbf{OK-VQA} \cite{marino2019ok} is a VQA dataset designed for knowledge-based VQA in natural scenes, which contains 14,055 questions and 14,031 images.
In OK-VQA, the visual content of the image is not sufficient to answer the question. A set of facts about natural scenes makes the connection between objects in the image and external knowledge text, which enables the prediction of answers. 
The retrieved knowledge text can serve as textual explanations for reasoning problems.

\textbf{KVQA} \cite{shah2019kvqa} is a dataset for the task of knowledge-aware VQA, consisting of 183K question-answer pairs involving more than 18K named entities and 24K images.
Every sample in KVQA requires multi-entity, multi-relation, and multi-hop reasoning over large Knowledge Graphs (KG) to achieve a prediction of VQA.
Compared to FVQA, the annotations of KG in KVQA attach importance to world knowledge rather than commonsense, enabling the identification of people, locations, and organizations in the text.
KVQA can be utilized by graph explanation methods.

\textbf{VCR} \cite{zellers2019recognition} is a dataset consisting of 290k multiple choice QA problems derived from 110k movie scenes. VCR is designed for the task called Visual Commonsense Reasoning. 
In this task, given an image, a list of regions, and a question, a model need to answer the question and provide a rationale explaining the reasoning process.
In VCR, options of multimodal explanations are annotated for a higher-order cognitive and commonsense understanding of the world depicted by the image.

\textbf{2D Minecraft} \cite{sun2019program} is a program-guided policy learning game inspired by Minecraft. 
Each task corresponds to a crafted object, with complex goals requiring the agent to craft intermediate ingredients and build tools.
The agent learns through pre-defined annotated programs that guide its actions and decision-making. These programs serve as instructions for the agent to navigate, gather resources, and craft objects in the simplified Minecraft environment, allowing it to improve its problem-solving skills.

\textbf{Mocheg} \cite{yao2023end} is a large-scale dataset for multimodal fact-checking and explanation. Mocheg consists of 15,601 claims, 33,880 textual paragraphs, and 12,112 images. 
Each claim is annotated with a truthfulness label and a ruling statement, which is supported by textual paragraphs and images as multimodal evidence.
Through Mocheg, the truthfulness of the claims is assessed by retrieving multimodal evidence, predicting a truthfulness label, and generating textual explanations to explain the reasoning process.

\textbf{ACT-X} \cite{park2018multimodal} is a dataset designed for action explanation, which aims to provide both visual and textual justifications for classification decisions in activity recognition tasks. 
The dataset consists of 18,030 images, with each image accompanied by three explanations. ACT-X is a valuable dataset for assessing the extent to which models align with human perception in terms of the evidence supporting classification decisions.
%
ACT-X allows researchers to evaluate and enhance the performance of CMR models in generating multimodal explanations for activity recognition tasks.

\textbf{ScienceQA} \cite{lu2022learn} is a dataset for science question answering . 
ScienceQA comprises 21,208 multimodal multiple-choice questions covering various science topics. Each question is annotated with corresponding lectures and explanations. The majority of questions are provided with grounded lectures and explanations in the form of Chain-of-Thought. 
In summary, ScienceQA aims to facilitate the development of models capable of generating coherent Chain-of-Thought explanations to interpret the multi-step reasoning process when making reasonable predictions.

\textbf{HatReD} \cite{hee2023decoding} is a multimodal dataset specifically containing hateful memes, along with the associated hateful contextual reasons. 
It serves as a valuable resource for a conditional generation task, which focuses on automatically generating the underlying reasons that explain hateful memes. 
The primary objective of HatReD is to facilitate the evaluation of fine-tuned Pretrained Language Models (PLMs) in their ability to generate explanations for hateful memes within a previously unseen dataset specifically centered around misogynous memes.

\textbf{WAX} \cite{liu2022wax} is a dataset for word association which is a paradigm for studying the human mental lexicon.
WAX provides a collection of words, word association graphs, and explanations for the edges, helping researchers explore and understand the intricate network of word associations in the human mind. 
Its annotated textual explanations for the relations between words can be particularly beneficial for advancing the development of models that generate textual explanations for the reasoning process.

We summarize the abovementioned datasets in Table \ref{tab:dataset}.

\begin{table}[ht]
\centering
\caption{Summary of cross-modal reasoning datasets with annotations for explanations.}
\renewcommand\arraystretch{1.2}
\begin{tabular}{|c|c|c|}
\hline
\textbf{Dataset} & \textbf{Cross-modal Reasoning Tasks}             & \textbf{Annotation for Explanations}              \\ \hline
Visual Genome \cite{krishna2017visual}    & Phrase Grounding, Visual Question Answering & \begin{tabular}[c]{@{}c@{}}Visual Objects, Scene Graphs, \\ Region Descriptions\end{tabular} \\ \hline
GQA \cite{hudson2019gqa}              & \multirow{6}{*}{Visual Question Answering}       & \begin{tabular}[c]{@{}c@{}}Visual Objects, Scene Graphs, \\ Functional Programs\end{tabular} \\ \cline{1-1} \cline{3-3} 
GQA-REX \cite{chen2022rex}          &                                                  & Multimodal Explanations                           \\ \cline{1-1} \cline{3-3} 
FVQA \cite{wang2017fvqa}             &                                                  & Supporting Facts                                  \\ \cline{1-1} \cline{3-3} 
VQA-E \cite{li2018vqa}            &                                                  & Image Captions Related to Questions               \\ \cline{1-1} \cline{3-3} 
OK-VQA \cite{zellers2019recognition}           &                                                  & External Knowledge Text                           \\ \cline{1-1} \cline{3-3} 
KVQA \cite{shah2019kvqa}             &                                                  & Knowledge Graphs                                   \\ \hline
VCR \cite{zellers2019recognition}              & Visual Commonsense Reasoning                     & Multimodal Rationale Options                      \\ \hline
2D Minecraft \cite{sun2019program}     & Program-guided Policy Learning                   & Programs                                          \\ \hline
Mocheg \cite{yao2023end}           & Multimodal Fact-checking and Explanation         & Multimodal Evidences                              \\ \hline
ACT-X \cite{park2018multimodal}            & Activity Recognition                             & Multimodal Action Explanations                    \\ \hline
ScienceQA \cite{lu2022learn}        & Science Question Answering                       & Lectures, Chain-of-Thought Explanations           \\ \hline
HatReD \cite{hee2023decoding}           & Hateful Meme Explanation                         & Textual Reasons                                   \\ \hline
WAX \cite{liu2022wax}              & Word Association                                 & Explanations for Associations                      \\ \hline
\end{tabular}
\label{tab:dataset}
\end{table}

\section{Challenges and Future Directions}
Although interpretable cross-modal reasoning (I-CMR) has achieved substantial advancements, especially in recent years, there still exist significant problems in the research.
In this section, we summarize the challenges existing in the research of I-CMR and provide possible future directions.

\subsection{Inaccurate Grounding of Visual Objects}
\label{sec:invb}
In CMR tasks involving visual inputs (e.g., images and videos), an important task of the visual explanation is grounding key visual objects for reasoning.
Existing explanation methods with visual object grounding can be roughly divided into two groups. 
The first category is grounding important grids for the reasoning results or relevant entities in another modality.
However, computing the importance map for the visual input is still a challenging problem \cite{selvaraju2017grad, srinivas2019full, jiang2021layercam}. Current methods usually focus on a limited region of the key object and its adjacent region, neglecting the global characteristics of the object.
The second category is using pretrained detection models (e.g., Faster-RCNN \cite{ren2015faster}) to pre-extract object boxes and then ground key object boxes for explanations.
However, lacking information regarding a specific reasoning problem, the pretrained detection model may fail to pre-extract the key objects for the reasoning problem.
Furthermore, the object box usually cannot precisely conform to the shape of the target object and may involve multiple objects simultaneously.
To sum up, the grounded visual objects by existing methods for explanation are often inaccurate.
Recently, Segment Anything Model (SAM) \cite{kirillov2023segment} has demonstrated remarkable proficiency in finely segmenting visual objects.
However, the segmentation results of SAM exhibit variability due to ambiguity and depend on some hyper-parameters, such as point prompts.
Therefore, accurately and finely grounding key visual objects by incorporating the semantics of a specific problem into SAM-based models may be a promising direction for visual explanations.

\subsection{Evaluation of Graph and Symbol Explanations}
While traditional CMR primarily focuses on evaluating the reasoning results, the evaluation of the provided explanations is also crucial for I-CMR to assess interpretability.
To evaluate the visual explanations, we can adopt widely-used metrics in computer vision tasks, such as object detection and semantic segmentation.
For example, Intersection over Union (IoU) \cite{padilla2020survey} is a measure that shows how well the prediction bounding box aligns with the ground truth box.
To evaluate the textual explanations, we can directly borrow metrics for natural language generation, such as BLEU \cite{papineni2002bleu}, METEOR \cite{banerjee2005meteor}, CIDEr \cite{vedantam2015cider}, and ROUGE \cite{lin2004rouge}.
Despite the relatively sufficient metrics for visual and textual explanations, the evaluation metrics for graph and symbol explanations are still under-explored.
To evaluate the quality of the generated graph explanations, there is currently a lack of universally accepted evaluation criteria for graph alignment.
Moreover, the evaluation of the symbol inference processes is also difficult due to the presence of multiple potential inference paths.
Therefore, developing metrics to evaluate the quality of the generated graphs and symbol inference processes may be a crucial direction for future research on graph and symbol explanations.

\subsection{Multimodal Explanation Annotations}
As also discussed in Section \ref{sec:mul}, multimodal explanations offer more comprehensive insights into the reasoning process and can better fit the inherent multimodal nature of CMR tasks.
Within the realm of multimodal explanation, the category of joint explanation combines multiple modalities to form a unified explanation, thereby further enhancing interpretability and user-friendliness compared to the category of independent explanation.
For example, REX \cite{chen2022rex} utilizes textual description and grounds key objects in the image to clearly explain the cross-modal reasoning process, as shown in Fig. \ref{fig:17-1}.
However, the annotation of multimodal explanations, particularly joint explanations, presents a more intricate and challenging task than single-modal explanations since the annotators have to utilize multiple modalities simultaneously in the annotation.
Currently, the CMR datasets containing annotations for multimodal explanations remain limited in quantity.
For example, GQA-REX \cite{chen2022rex} generates joint visual-and-textual explanations for visual question answering, which uses pretrained Faster-RCNN to extract visual objects.
Subsequently, the annotated visual objects in explanations are often inaccurate, which is also discussed in Section \ref{sec:invb}.
Therefore, despite the promising interpretability of multimodal explanations, the construction of CMR datasets with high-quality multimodal explanations still necessitates substantial efforts from researchers. 
This is crucial in order to facilitate further advancements and research in the field of multimodal explanation for CMR.

\subsection{User-friendliness and User-interaction}
I-CMR has demonstrated significant application values in various domains especially those that heavily rely on reliability and security, including transportation, finance, and healthcare \cite{sachan2021evidential, zhan2022uniclam, wang2023decision}.
In real-world applications, a crucial concern is facilitating users to understand the provided explanations for the CMR process.
Consequently, the users can make the decision to accept or reject the reasoning result of a specific CMR problem based on the provided explanation.
However, some explanations offered by current methods may be hard to understand for general users lacking expert knowledge of machine learning, or may require an excessive amount of time to comprehend. 
For example, X-Pool \cite{gorti2022x} visualizes the relevance between the query text and all video frames for text-video retrieval, where the relevance to most frames is redundant for a specific problem.
Moreover, current methods mostly provide the explanation in one shot.
Nevertheless, the users may be still confused with the explanation and require further clarification.
Existing explanation methods for CMR lack the ability to provide additional explanations to address users' specific confusion.
Therefore, improving the user-friendliness of the provided explanations and implementing dynamic explanations through user interactions can be promising directions in the research of I-CMR.

\subsection{Comprehensive Explanation Based on Large Language Models}
In the field of I-CMR, comprehensively explaining the reasoning process is still a key challenge.
Despite the multimodal nature of the input, some methods (e.g., FiLM \cite{perez2018film}, TAA \cite{li2018tell}, and NPN \cite{zhou2017more}) merely aim to interpret one modality of the input, which is insufficient to explain the overall reasoning process.
Moreover, some methods (e.g., VRANet \cite{yu2020reasoning}, CKRM \cite{wen2020multi}, CRRN \cite{chen2021cross}) merely provide only limited insights into the reasoning process, such as locating the important regions in inputs, retrieving relevant knowledge, and predicting the relations between regions, which is far from completely explaining the reasoning process.
Since Large Language Models (LLMs) \cite{brown2020language, du2022glm, touvron2023llama} have recently shown unparalleled capabilities of natural language comprehension and generation, there has been a growing interest in generating comprehensive explanations of reasoning processes based on LLMs \cite{lu2022learn, wang2023t}.
%
However, developing multimodal LLMs for cross-modal reasoning tasks is still an open problem.
The architecture of the LLM should be capable of processing and integrating information from multiple modalities.
Moreover, generating explanations of modalities other than text based on LLMs is also under-explored.
The generated explanations should reflect how the model leverages information from different modalities to arrive at its decision.
Overall, inspired by the recent advancements in the area of LLM, generating comprehensive explanations of reasoning processes based on LLMs is a promising and challenging research direction for I-CMR.

\section{Conclusions}
In recent years, there has been a growing interest in the interpretability of cross-modal reasoning (CMR), with a focus on moving beyond traditional black-box models and explaining the reasoning process to users.
This survey aims to provide a comprehensive overview of the work in the field of interpretable cross-modal reasoning (I-CMR).
First, we establish a unified three-level taxonomy for the existing methods of I-CMR.
Subsequently, we introduce typical methods for each category and analyze their similarities and limitations.
Moreover, we also introduce and categorize the current CMR datasets that include annotations for explanations.
However, there is still significant potential for advancement in I-CMR.
Therefore, we summarize the existing challenges and discuss future directions for I-CMR.
We hope this survey can furnish researchers interested in I-CMR with a comprehensive understanding and promote further progress in this field.


\bibliographystyle{ACM-Reference-Format}
\bibliography{bib}

\end{document}